\documentclass[runningheads]{llncs}

\usepackage{eccv}

\usepackage{bm}
\usepackage{eccvabbrv}

\usepackage{graphicx}
\usepackage{booktabs}
\usepackage{mathtools}
\usepackage{wrapfig}
\usepackage[accsupp]{axessibility}  
\DeclarePairedDelimiter\abs{\lvert}{\rvert}%
\DeclarePairedDelimiter\norm{\lVert}{\rVert}%

\makeatletter
\let\oldabs\abs
\def\abs{\@ifstar{\oldabs}{\oldabs*}}
\let\oldnorm\norm
\def\norm{\@ifstar{\oldnorm}{\oldnorm*}}
\makeatother

\usepackage{array}
\newcolumntype{P}[1]{>{\centering\arraybackslash}p{#1}}
\newcolumntype{M}[1]{>{\centering\arraybackslash}m{#1}}

\usepackage{hyperref}

\usepackage{orcidlink}

\begin{document}

\title{CoPT: Unsupervised Domain Adaptive Segmentation using Domain-Agnostic Text Embeddings} 

\titlerunning{CoPT}

\author{Cristina Mata \and 
Kanchana Ranasinghe \and 
Michael S. Ryoo} 

\authorrunning{C. Mata et al.}

\institute{Stony Brook University, Stony Brook, NY, USA \\
\email{\{cfmata, kranasinghe, mryoo\}@cs.stonybrook.edu}
\vspace{-1em}
}

\maketitle

\begin{abstract}
Unsupervised domain adaptation (UDA) involves learning class semantics from labeled data within a source domain that generalize to an unseen target domain. UDA methods are particularly impactful for semantic segmentation, where annotations are more difficult to collect than in image classification. Despite recent advances in large-scale vision-language representation learning, UDA methods for segmentation have not taken advantage of the domain-agnostic properties of text. To address this, we present a novel \textbf{Co}variance-based \textbf{P}ixel-\textbf{T}ext loss, CoPT, that uses domain-agnostic text embeddings to learn domain-invariant features in an image segmentation encoder. The text embeddings are generated through our LLM Domain Template process, where an LLM is used to generate source and target domain descriptions that are fed to a frozen CLIP model and combined. In experiments on four benchmarks we show that a model trained using CoPT achieves the new state of the art performance on UDA for segmentation. The code can be found at \url{https://github.com/cfmata/CoPT}.
\keywords{Unsupervised Domain Adaptation \and Semantic Segmentation \and Vision Language Models}
\end{abstract}

\section{Introduction}
\label{sec:introduction}
\begin{figure}[tb]
  \centering
  \includegraphics[height=5cm]{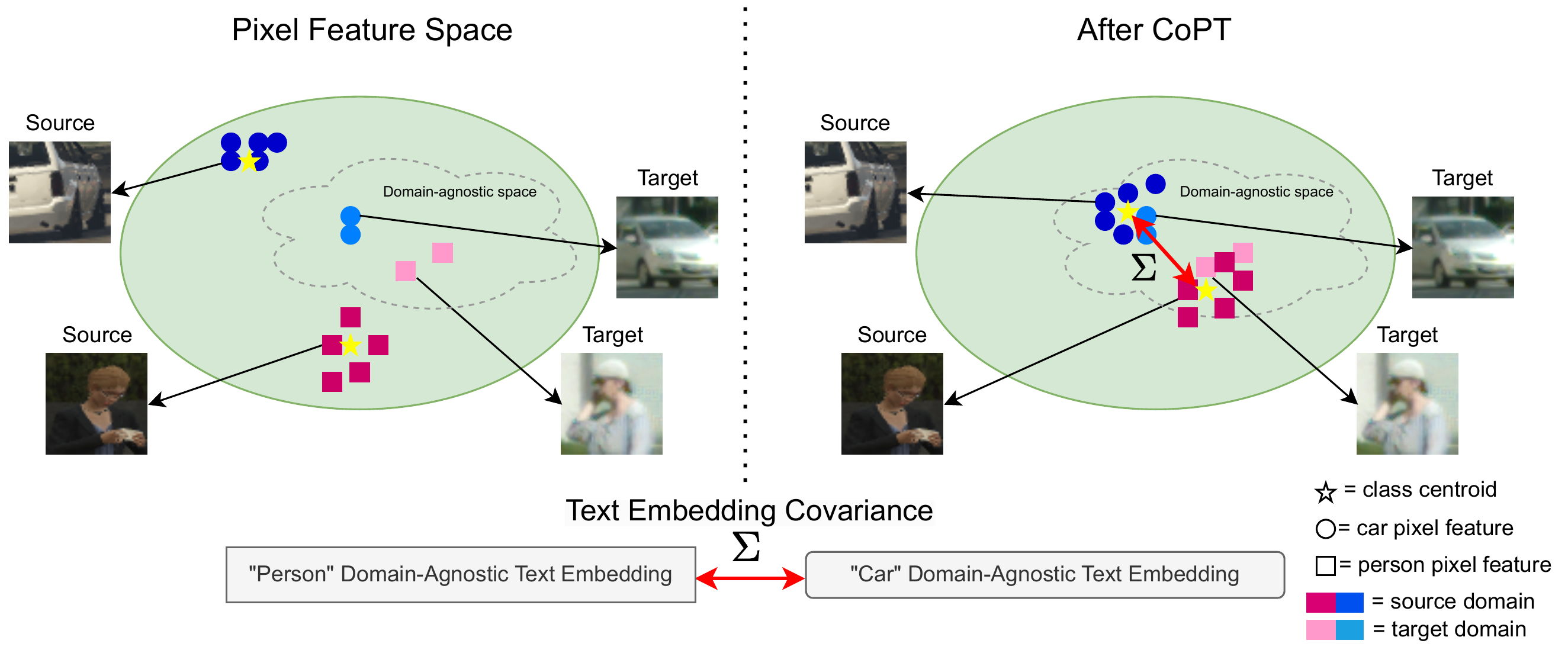}
  \vspace{-0.7em}
  \caption{CoPT encourages the distance between class centroids in pixel feature space to be the same as in domain-agnostic text embedding space.
  }
  \vspace{-1.1em}
  \label{fig:concept}
\end{figure}
\par Semantic segmentation models are used in safety-critical applications, such as surgical robotics and autonomous vehicles, where it is of the utmost importance to properly segment small, rare and moving objects. The pixel-wise classification task requires a copious amount of labeled training data to perform well, especially on small objects, but human-collected annotations are difficult to obtain at scale. In addition, annotations collected automatically using large internet-scale pre-trained models such as Segment Anything Model (SAM) \cite{Kirillov2023} are often noisy or lack semantic class associations, resulting in downstream errors. To avoid issues caused by imperfect labels, practitioners still train on synthetic data that emulates the target domain, or on real labeled data in a different source domain, before adapting the model to the target domain. The challenge remains to develop an unsupervised domain adaptation (UDA) method that learns strong semantic representations from available labels in a source domain while generalizing to the unlabeled target domain.
\par The dominant approach taken by UDA methods is self-training in a teacher-student paradigm \cite{Hoyer23, Hoyer22eccv, Hoyer22cvpr}, where an EMA-updated image encoder teacher produces pseudo-labels on target images for training the student, essentially bootstrapping unlabeled data with noisy labels to learn target domain features gradually. Self-training is an attractive approach due to its straightforward implementation and impressive results, but remains susceptible to noisy predictions from the teacher and dependent on the strength of the pretrained image encoder upon initialization. An orthogonal line of work is data augmentation \cite{Hoyer22cvpr}, which aims to make the model robust to changes in an class' appearance by augmenting the source samples directly with transformations. Data augmentation cannot simulate the true structure of objects in target domains, which inherently brings limited improvements to performance. Previous approaches do not target domain-specific features in latent representations learned by the image encoder. 
\par We propose a novel \textbf{Co}variance-based \textbf{P}ixel-\textbf{T}ext loss, CoPT, which learns domain-agnostic representations for objects in images using domain-agnostic representations from a different modality: text. Text embeddings of class names are domain-agnostic \cite{Bose24} because they do not contain the visual domain-specific bias that prevents source-supervised segmentation networks from performing well on the target domain. The word for ``cat'' remains the same regardless of the animal's appearance in a real or synthetic image. It follows that text can be used as a regularization tool to encourage learning domain-invariant features in visual embeddings. This idea was developed and validated by RISE \cite{Huang23} for image classification, but in that work both text and visual embeddings came from the same CLIP \cite{Radford21} model. CoPT claims that this type of regularization works even when text and visual embeddings lie in separate latent spaces and backs this claim with strong empirical results (Section \ref{sec:results}).
\par To guide learning domain-invariant pixel features in a segmentation encoder, CoPT enforces the distance between pixel features from different classes to be the same in a segmentation encoder's latent space as it is in a text encoder's latent space. To circumvent the hindrance that pixel and text features lie in different latent spaces, CoPT minimizes the distance between the covariance matrix of pixel and text features. Figure \ref{fig:concept} conceptually shows how CoPT is guided by domain-agnostic text embeddings to find domain-agnostic pixel features. Rather than relying on hand-crafted text templates to generate domain-agnostic text embeddings, we propose LLM Domain Template, a process through which an LLM (Large Language Model) is used to describe domain properties.
\par We validate CoPT on four UDA segmentation benchmarks and show that our method surpasses the  current state of the art. We perform ablations on GTA$\rightarrow$Cityscapes to show that CoPT can boost the performance of multiple UDA methods and to show the benefits of domain-agnostic text embeddings generated from LLM Domain Template over hand-crafted ones. Our main contributions are,
\begin{enumerate}
    \item The first method to leverage vision-language representations in UDA for semantic segmentation.
    \item LLM Domain Template, a framework for using LLMs to generate domain-agnostic text embeddings.
    \item CoPT, a novel covariance-based pixel-text loss that bridges separate image and text latent spaces in order to learn domain-agnostic features.
\end{enumerate}
\begin{figure}[tb]
  \centering
  \includegraphics[width=\textwidth]{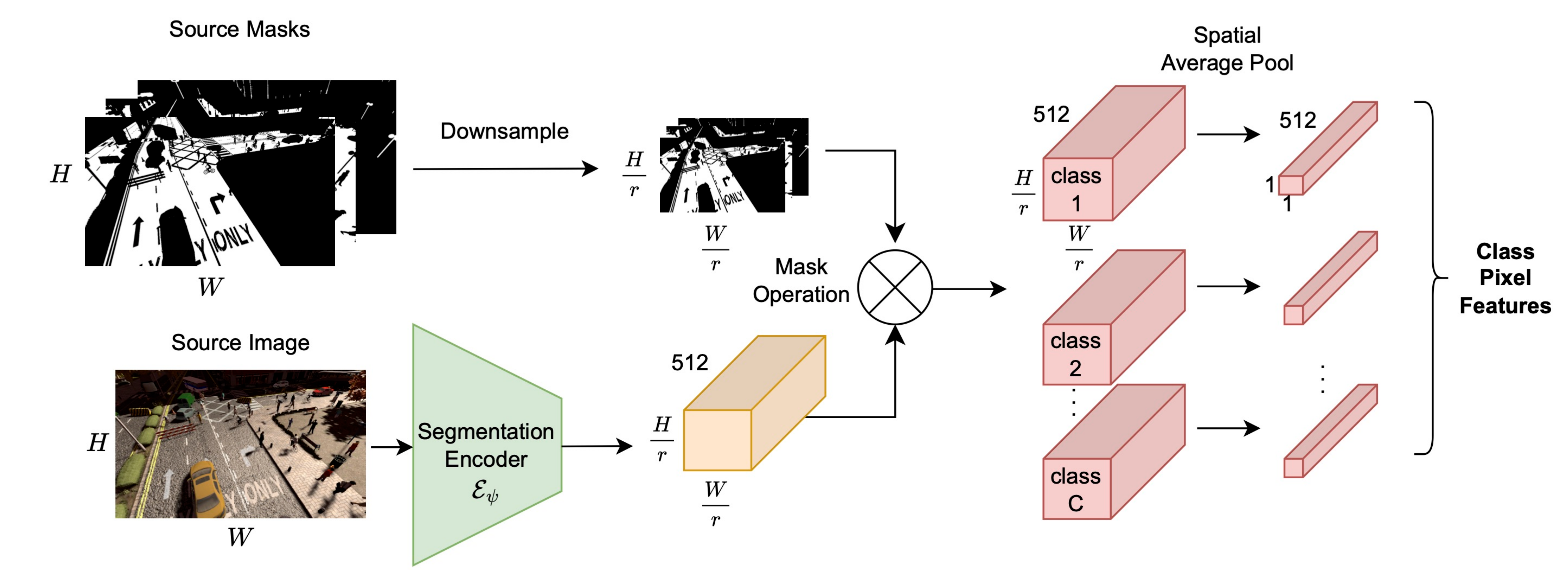}
   \vspace{-1.0em}
  \caption{Class pixel features are generated by first extracting $\frac{H}{r}\times\frac{W}{r}\times512$ resolution feature maps from source images using $\mathcal{E}_{\psi}$, where $r$ is the encoder's spatial downsampling factor. Each class' binary ground truth mask is downsampled to the same resolution. The class mask is applied to the low resolution feature, then spatial averaging is used to generate a $1\times1\times512$ feature for each class.
  }
  \vspace{-1.0em}
  \label{fig:pixel_feature_generation}
\end{figure}
\section{Related Work}
\label{sec:related_work}

\subsection{Unsupervised Domain Adaptation for Semantic Segmentation}
\label{sec:uda}
\par Unsupervised domain adaptation (UDA) for semantic segmentation methods can be categorized through their use of four strategies: adversarial learning, data augmentation, self-supervision and prior learning. Self-training in particular has been shown to give large boosts in performance and many variations of it have been explored \cite{Hoyer23,Hoyer22eccv,Chen22,Lu22, Zhang21,Wang22,Kundu21, Li20, Zheng20}.
While this approach strengthens representations of common classes, the predictions can be noisy and the model suffers on small and rare classes. Data augmentation strategies in the form of strongly augmented self-training \cite{Hoyer22cvpr,Hoyer22eccv} and source-target image/label mixing \cite{Chen23pipa} has brought strong improvements. 
\par Our work presents a major departure from previous approaches. To the best of our knowledge we present the first work to incorporate vision-language representations into UDA for semantic segmentation. While this idea has been studied for image classification \cite{Huang23}, it is not directly transferable to segmentation representations because of the need to conserve spatial resolution in the encoder. A comprehensive study of the best ways to use vision-language representation learning for UDA in segmentation is needed. We hope our work will encourage further methods to take advantage of recent advances in vision-language representation learning and the domain-agnostic properties of text.
\subsection{Domain Adaptation using Vision-Language Embeddings}
\label{sec:da_vl}
\par Since the introduction of CLIP \cite{Radford21}, domain adaptation methods have taken advantage of its rich vision-language latent space in chiefly two ways: 1. adapting the image embedding space to be more similar to the text embedding space \cite{Lai23} or 2. prompt tuning \cite{Lai24,Zara23,Bose24,Cao23}. Unfreezing CLIP for target adaptation \cite{Lai23} is computationally expensive and risks harming CLIP's strong zero-shot generalizability. Instead of directly fine-tuning CLIP, methods that take the prompt tuning approach capitalize on its generalizability by embedding the text prompts describing the target domain into a space better suited to the target domain. Our method is most inspired by RISE \cite{Huang23} which generates domain-agnostic text embeddings and uses these to search for domain-agnostic image representations. The key difference lies in how we generate text descriptions for our domains: while their approach uses simple hand-crafted domain descriptors, we develop LLM Domain Template, a process to generate descriptions of each domain. In addition, their approach is designed for image classification, allowing them to use CLIP's image encoder directly for downstream classification, whereas we are focusing on segmentation and must grapple with anchoring pixel-specific features to text embeddings from a separately learned space. Jin et al. \cite{Jin24} similarly use domain descriptors for domain adaptive object detection, a task that also requires feature localization, but their method promotes visual context learning instead of using domain-agnostic text embeddings.
A comprehensive review of related works can be found in Appendix \ref{sec:appendix_related_works}.

\section{Methods}
\label{sec:methods}

%

\subsection{Preliminaries}
\label{sec:preliminaries}

\par In unsupervised domain adaptation (UDA) for semantic segmentation, we are given access to a labeled source dataset $S = \{(\bm{x}_{i}, \bm{y}_{i})\}_{i=1}^{N_{S}}$ of $N_{S}$ image and mask pairs $(\bm{x},\bm{y})$ that have spatial height and width $(H,W)$. Every pixel in the ground truth mask $\bm{y}$ falls into one of $C$ classes: $\{\bm{y} \in \mathbb{Z} : 0 \leq \bm{y} \leq C\}$. The goal is to learn a model $\mathcal{F}_{\theta}$ parameterized by $\theta$ that performs pixel-wise classification well on inputs from an unlabeled target dataset $T = \{\bm{x}_{i}\}_{i=1}^{N_{T}}$.
\par Typically, the segmentation network $\mathcal{F}_{\theta}$ is trained to minimize the following pixel-wise cross entropy loss $\mathcal{L}_{ce}$ over each source image sample \cite{Zheng20,Tsai18,Li20},
\begin{equation}
\mathcal{L}_{ce} = - \sum \limits_{i=1}^{H} \sum \limits_{j=1}^{W} \sum \limits_{c=1}^{C} \mathbb{I}[y_{ijc} = c] \text{log} (\hat{y}_{ijc})
\label{eq:ce_loss}
\end{equation}
where $\hat{\bm{y}} = \mathcal{F}_{\theta}(\bm{x})$ is the model's predicted discrete distribution over all classes $C$ for the image $\bm{x}$ and $\mathbb{I}[y_{ijc} = c]$ is a binary indicator for whether the pixel at $(i,j)$ belongs to class $c$. Additionally, many UDA methods optimize auxiliary losses to modify $\mathcal{F}_{\theta}$'s latent space \cite{Hoyer23,Chen23pipa}. 
\par In our approach, we introduce a loss that modifies the distance between $\mathcal{F}_{\theta}$'s class representations to be the same as in the latent space of a frozen CLIP-pretrained text encoder $\mathcal{G}$. To the best of our knowledge, this is the first UDA work to apply vision-language representations for semantic segmentation. We first build domain-agnostic text embeddings by averaging embeddings of domain-specific prompts generated by a Large Language Model in our LLM Domain Template process (Section \ref{sec:llm_templates}). Then we extract pixel-wise class features from the image segmentation encoder (Figure \ref{fig:pixel_feature_generation}). Lastly our proposed covariance-based loss CoPT bridges the gap between the image encoder and text encoder latent spaces in order to enforce learning of domain-agnostic features (Section \ref{sec:copt}).

\subsection{Hand-Crafted Domain-Agnostic Text Embeddings}
\label{sec:text_generation_handcraft}

\par We have access to a CLIP-pretrained text encoder $\mathcal{G}$ that takes a string prompt $p$ and outputs a text embedding $\bm{t} = \mathcal{G}(p)$. In our work we keep $\mathcal{G}$ frozen and use it purely to extract text embeddings. The prompt template specific to domain $d$ and class $c$ is hand-crafted in \cite{Huang23} as, 
\begin{align}
    p_{d}(c) = \text{"A <DOMAIN> of a <CLASS>"}
\end{align}
where $d = \text{DOMAIN}$ and $c=\text{CLASS}$. For example, when the target domain is photos of objects, one would plug in ``photo'' to $\text{DOMAIN}$ so that ``A photo of a car'' denotes the target domain description of a car. 

\begin{wrapfigure}[15]{R}{0.5\textwidth}
\centering
\vspace{-2em}
\includegraphics[width=6cm]{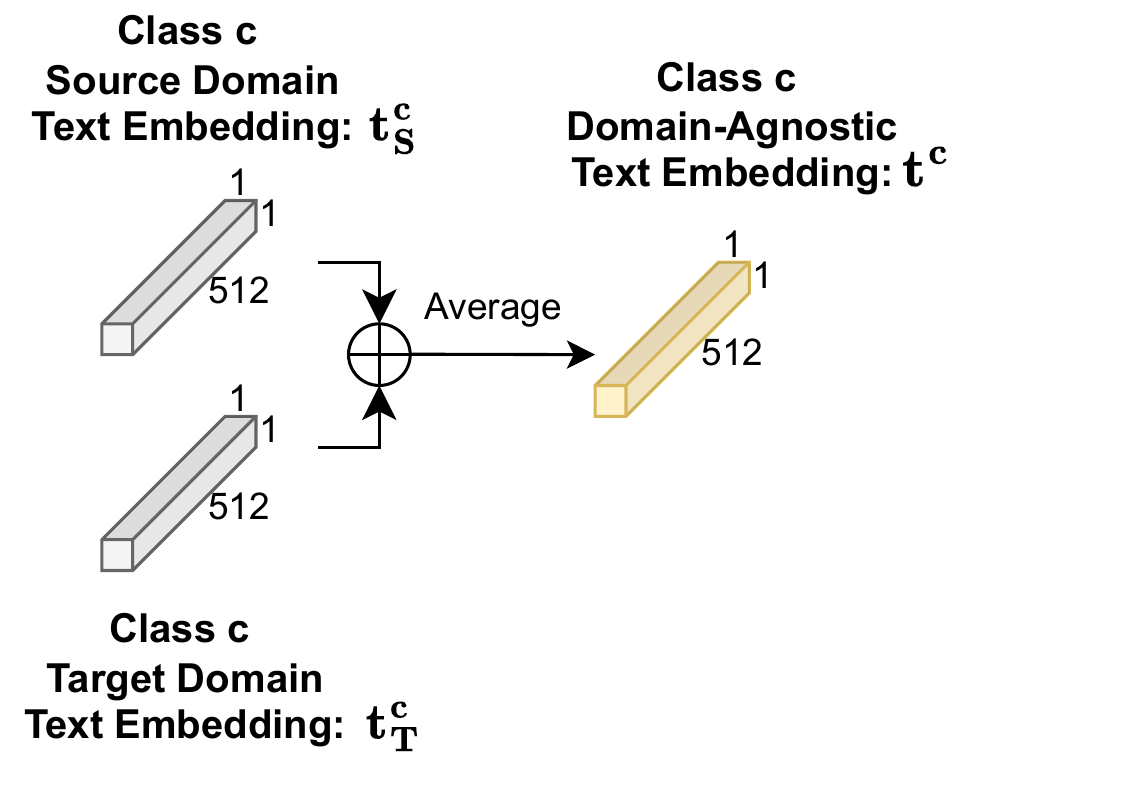}
\vspace{-2em}
\caption{The final domain-agnostic text embedding $\mathbf{t^{c}}$ for class $c$ is generated by averaging the source and target text embeddings, $\mathbf{f_{S}^{c}}$ and $\mathbf{f_{T}^{c}}$.}
\label{fig:source_target_text_comb}
\end{wrapfigure}

\par A prompt template is crafted for each domain so that, given a domain $d$ and a class $c$, the text embedding $\bm{t}_{d}^{c} = \mathcal{G}(p_{d}(c))$ contains domain-specific and class-specific information. According to RISE \cite{Huang23}, simply averaging the text embeddings of multiple class templates results in a generic text representation of an object. This ensures that the text embedding is representative of a class across multiple domains. To generate a domain-agnostic prompt $\bm{t}^{c}$ for a class $c$, the source and target embeddings are averaged together as shown in Figure \ref{fig:source_target_text_comb}:
\begin{align}
\bm{t}^{c} = \frac{1}{2} (\bm{t}_{S}^{c} + \bm{t}_{T}^{c})
\label{eqn:source_target_average}
\end{align}

\noindent Such a generalized text embedding of a class is called domain-agnostic text embedding by Bose et al. \cite{Bose24} and we follow this terminology. While we follow RISE's hand-crafted prompt construction method, we modify it to use two domain templates (for source and target) instead of the 80 CLIP templates \cite{Radford21}.
\par So far, prior works have relied on using hand-crafted domain descriptors to embed semantic information specific to a domain. However these descriptors are short and lack details about what contributes to the domain gap, namely the texture and appearance of a domain. We propose a framework called LLM Domain Template to increase the semantic expressiveness of the text embeddings.

\subsection{LLM Domain Template}
\label{sec:llm_templates}
\par
\begin{figure}[tb]
  \centering
    \includegraphics[width=12.2cm]{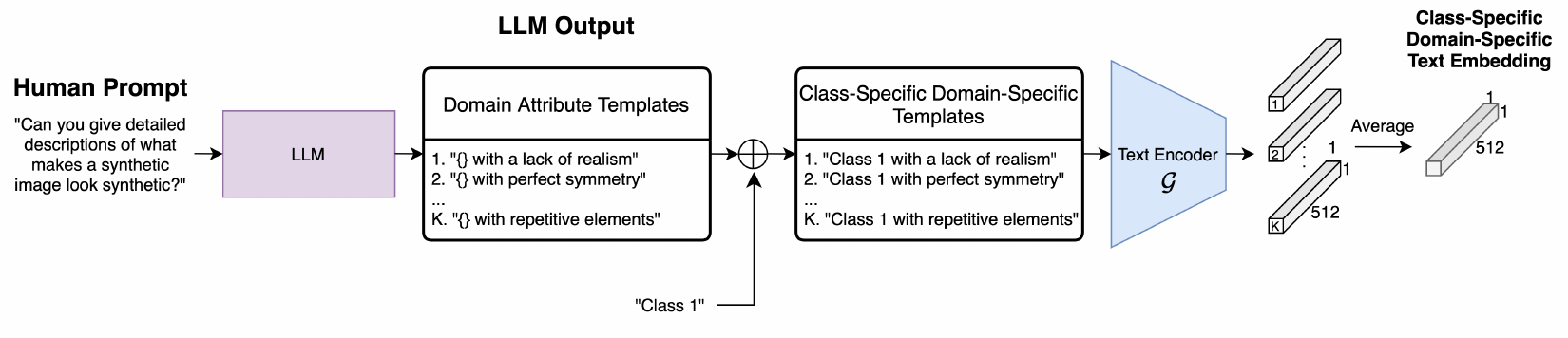}
    \caption{LLM Domain Template: An LLM is queried to describe the source and target domain attributes. Here the query is about the synthetic source domain. The resulting templates are formatted with a class name, fed to a frozen CLIP text encoder and averaged to obtain a single text embedding for a class.}
    \label{fig:text_embedding_generation}
\end{figure}

 LLM Domain Template uses a large language model, ChatGPT \cite{openai2024gpt4}, to automatically generate long-form descriptors of the source and target domains. We hypothesize that doing so will lead to better domain-agnostic text embeddings because the average embedding will be closer to the centroid of the domain-specific regions in latent space. Figure \ref{fig:text_embedding_generation} shows LLM Domain Template in action. We query ChatGPT with "Can you give detailed descriptions of what makes a <DOMAIN> image look <DOMAIN>?" and get a list of $K$ domain attribute templates. The prompt template specific to attribute $k = \text{ATTRIBUTE K}$ generated by ChatGPT is
\begin{align}
    p_{k}(c) = \text{"A <CLASS> with <ATTRIBUTE K>"}
    \label{eq:llm_query}
\end{align}
We fill in the class name and attribute for all templates, then feed them to the text encoder $\mathcal{G}$ and average them to get the domain-specific text embedding for a class:
\begin{align}
    \bm{t}_{d}^{c} = \frac{1}{K} \sum \limits_{k=1}^{K} \mathcal{G}(p_{k}(c))
\end{align}
\par The final domain-agnostic text embedding for a class, $\bm{t}^{c}$, is generated using Equation \ref{eqn:source_target_average}. The domain-agnostic embeddings are generated once at model initialization and stored in a memory bank.

\subsection{CoPT: Covariance-based Pixel-Text Loss}
\label{sec:copt}

\par Our goal is to adapt the segmentation encoder's latent space to the target by learning domain-agnostic class representations. We use the domain-agnostic text embeddings $\bm{t}^{c}$ to structure this representation space using a covariance-based pixel-text loss, or CoPT. CoPT enforces the distance between classes in pixel feature space to be the same as the distance between classes in text embedding space. This requires extracting class pixel features from the encoder given a source domain sample.
\par Our image segmentation model $\mathcal{F}_{\theta}$ is composed of an encoder $\mathcal{E}_{\psi}$ and decoder $\mathcal{D}_{\rho}$: $\mathcal{F}_{\theta} = \mathcal{D}_{\rho}(\mathcal{E}_{\psi})$. $\mathcal{E}_{\psi}$ spatially downsamples its input by a factor $r$, so while input image $\bm{x}$ has resolution $(H,W)$, the encoder's features $\bm{f} = \mathcal{E}_{\psi}(\bm{x})$ have resolution $(\frac{H}{r}, \frac{W}{r})$. To extract class pixel features from $\bm{f}$, we spatially downsample $\bm{x}$'s corresponding ground truth mask $\bm{y}$ to $(\frac{H}{r}, \frac{W}{r})$ and generate a binary mask $\bm{b}^{c}$ for class $c$ with the same dimensions:
\begin{align}
    \bm{b}^{c} = \mathbb{I}[\text{downsample}(\bm{y}) = c]
\end{align}
The mask $\bm{b}^{c}$ is multiplied by $\bm{f}$ to extract class features and spatial averaging is applied to get the final pixel feature for a class, $\bm{f}^{c}$:
\begin{align}
    \bm{f}^{c} = \frac{1}{(\frac{H}{r})(\frac{W}{r})} \sum \limits_{i=1}^{\frac{H}{r}} \sum \limits_{j=1}^{\frac{W}{r}} b_{i,j}^{c} \bm{f}_{i,j}
\end{align}
where $\bm{f}_{i,j}$ is the pixel feature at location $(i,j)$ and $b_{i,j}^{c}$ is a binary value indicating whether the pixel belongs to class $c$. Figure \ref{fig:pixel_feature_generation} illustrates the class pixel feature extraction process.
\par The pixel covariance matrix $\Sigma_{p}$ for an input sample $\bm{x}$ is calculated by measuring the cosine similarity between all pairs of class pixel features $\bm{f}^{c}$ from $\bm{x}$. A similar process is carried out with the domain-agnostic text embeddings $\bm{t}^{c}$ to get the text covariance matrics $\Sigma_{t}$:
\begin{align}
    \Sigma_{p} = \begin{bmatrix}
\text{Var}(\bm{f}^{1}) & \hdots & \text{Cov}(\bm{f}^{1}, \bm{f}^{C})\\
\vdots & \ddots & \vdots \\
\text{Cov}(\bm{f}^{C}, \bm{f}^{1}) & \hdots & \text{Var}(\bm{f}^{C})
\end{bmatrix}
,
\Sigma_{t} = \begin{bmatrix}
\text{Var}(\bm{t}^{1}) & \hdots & \text{Cov}(\bm{t}^{1}, \bm{t}^{C})\\
\vdots & \ddots & \vdots \\
\text{Cov}(\bm{t}^{C}, \bm{t}^{1}) & \hdots & \text{Var}(\bm{t}^{C})
\end{bmatrix}
\end{align}

The best segmentation performance is usually achieved by training with large input resolutions, leading to low batch sizes in order to fit images into GPU memory. As a result it is rare that all classes appear in a mini-batch. To address this we construct the text embedding's covariance matrix $\Sigma_{t}$ for corresponding classes as they occur in a sample during training. Low batch sizes also lead to high variance in class pixel features that jeopardizes stable training. To reduce variance in gradient updates we store the average pixel feature for each class in a memory bank, and update the bank on each iteration using a linear decay. That is, upon calculation of a pixel feature $\mathbf{f^{c}}$ for a class, the class in memory $\mathbf{f^{c}_{\text{current}}}$ is updated with decay $\lambda$ following 
\begin{align}
\mathbf{f^{c}_{\text{new}}} = \lambda * \mathbf{f^{c}_{\text{current}}} + (1-\lambda) * \mathbf{f^{c}}
\label{eq:membank_update}
\end{align}
When calculating the pixel covariance matrix, the updated pixel feature in the memory bank $\mathbf{f^{c}_{\text{new}}}$ is used. 
\par CoPT's final objective is to minimize the cosine distance between the covariance matrices $\Sigma_{p}$ and $\Sigma_{t}$:
\begin{align}
    \mathcal{L}_{\text{CoPT}} = 1 - \frac{\Sigma_{p} \cdot \Sigma_{t}}{\norm{\Sigma_{p}} \cdot \norm{\Sigma_{t}}}
\end{align}
\par
\begin{figure}[tb]
  \centering
  \includegraphics[height=5cm]{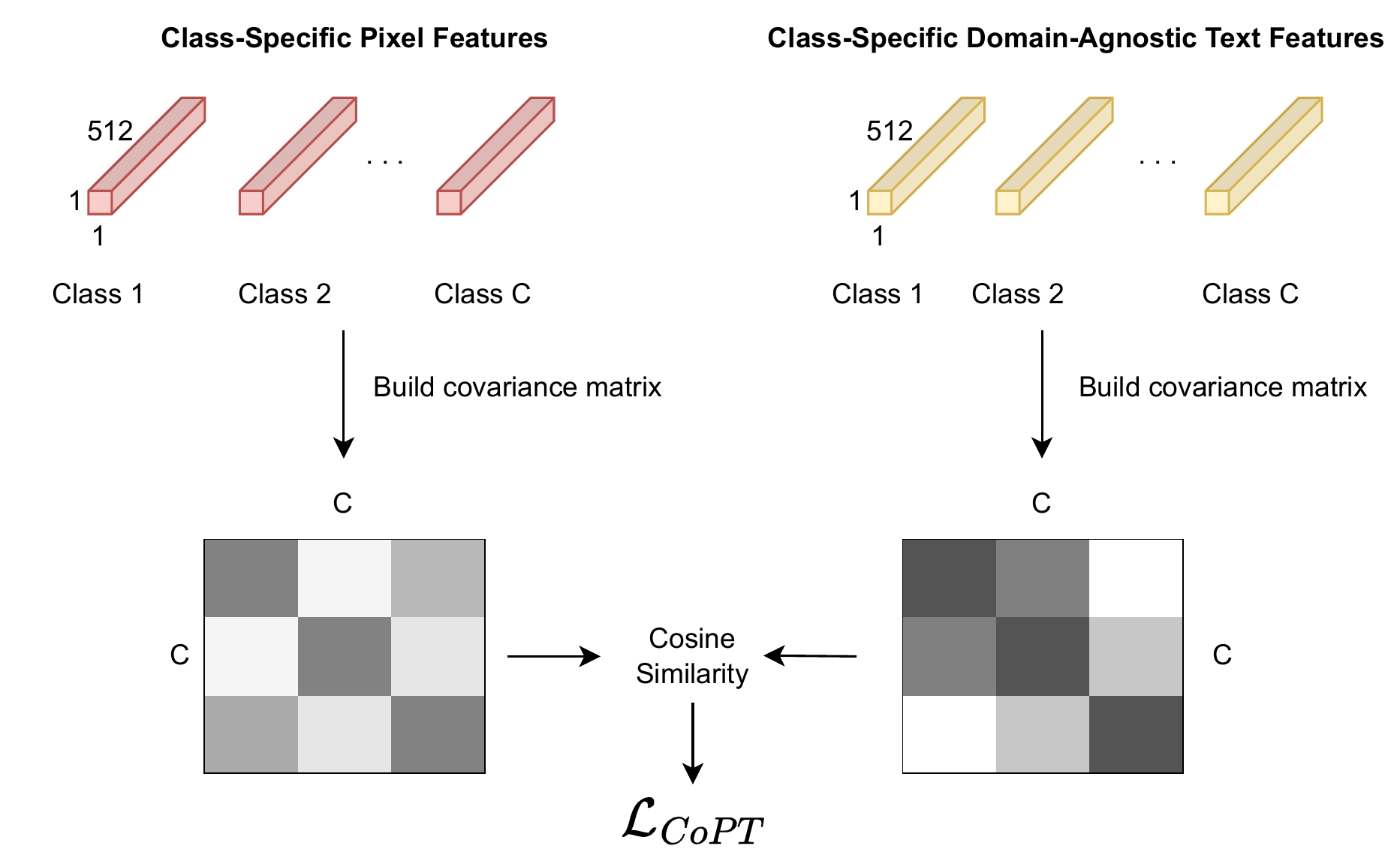}
  \caption{CoPT calculates the difference in pixel and text covariance matrices in order to transform the segmentation encoder's latent space to emulate the domain-agnostic text space.
  }
  \label{fig:covariance_calculation}
\end{figure}

 This calculation is illustrated in Figure \ref{fig:covariance_calculation}. CoPT can be added to any UDA training scheme that uses source images and their labels. 

\section{Experiments}
\label{sec:experiments}

\subsection{Datasets}
\label{sec:datasets}
\par We run experiments on four benchmarks commonly used for UDA for semantic segmentation: GTA$\rightarrow$CS \cite{Richter16}, Synthia$\rightarrow$CS \cite{Ros16}, CS$\rightarrow$DZ \cite{Sakaridis19} and CS$\rightarrow$ACDC \cite{Sakaridis21}. GTA$\rightarrow$CS and Synthia$\rightarrow$CS consider synthetic images as the source domain and real images as the target. This emulates the common industry practice of developing in-house synthetic datasets for the source domain. CS$\rightarrow$DZ considers the adaptation of day-time source to night-time target images. CS$\rightarrow$ACDC considers adaptation of clear to adverse weather conditions. All benchmarks are of road scenes, which is grounded in the importance of segmentation for autonomous vehicle applications. GTA \cite{Richter16} consists of 24,966 synthetic images of street scenes collected from a video game and comes with pixel-level semantic labels for 19 classes. Cityscapes (CS) \cite{Cordts16} is a real street-scene dataset with fine pixel-level annotations across the same 19 classes as GTA for 2,975 training images and 500 validation images. DarkZurich (DZ) \cite{Sakaridis19} consists of 2,416 training images of twilight scenes and 151 testing images, and has labels for the 19 CS classes. ACDC \cite{Sakaridis21} consists of 1600 training, 406 validation and 2000 test images and is labeled with the 19 CS classes. Synthia \cite{Ros16} consists of 9,400 synthetic images collected in a virtual city and contains annotations for 16 of the 19 CS classes. GTA$\rightarrow$CS and Synthia$\rightarrow$CS use the 500 labeled images from the CS validation set for evaluation. The DZ training set is used as an unlabeled target image source while the 151 images in the testing set are used only for evaluation. Similarly we use the ACDC training set as an unlabeled target image source and evaluate on its 2000 test images. In our experiments we report per-class \% Intersection over Union (IOU) as well as the average mIOU across 19 classes in GTA$\rightarrow$CS, CS$\rightarrow$DZ and CS$\rightarrow$ACDC, and across 16 classes in Synthia$\rightarrow$CS.

\subsection{Implementation}
\label{sec:implementation}
\par For LLM Domain Template, we query ChatGPT \cite{openai2024gpt4}, a transformer-based LLM trained for next-token-prediction on public internet and private licensed data. We submit one query for each domain following Equation \ref{eq:llm_query}. A full list of the query and responses can be found in Appendix \ref{sec:supp_llm_attributes}. Once templates are formatted with class names, they are passed to a frozen ViT-B/32 CLIP \cite{Radford21} text encoder to generate text embeddings, which takes 5.6 seconds on an NVIDIA RTX A5000. For our memory bank we use a dictionary initialized with empty Tensors of length 512. We perform an ablation experiment, detailed in section \ref{sec:supp_membank_ablation} to find the optimal decay $\lambda$ is 0.5. Our implementation is in PyTorch. The code may be found at \url{https://github.com/cfmata/CoPT}.
\par In our main experiments we implement CoPT on top of MIC \cite{Hoyer23}, the current state of the art for UDA for semantic segmentation. In MIC, the auxiliary losses are pixel-wise cross entropy over source images, masked pseudo-label self-training using an EMA teacher, ImageNet feature distance regularization, and strongly augmented self-training. CoPT is added with weight 1 to the auxiliary losses. We include detailed equations for self-training losses in Appendix \ref{sec:auxiliary_losses} but strongly encourage the reader to refer to \cite{Hoyer23} for complete details on the auxiliary losses and their hyperparameters. CoPT is optimized by Adam with learning rate 0.00006, betas (0.9, 0.999) and weight decay 0.01. We use batch size 2 per GPU and train for 40,000 iterations, which takes approximately 34 hours on a single NVIDIA RTX A5000. Image crops sized $1024\times1024$ are fed to the model during training.

\section{Results}
\label{sec:results}

\subsection{Unsupervised Domain Adaptation for Semantic Segmentation}

\par
\begin{figure*}[t]
\tabcolsep 0.03cm
\noindent\makebox[\textwidth]{
\begin{tabular}{ccccc}
  \includegraphics[trim = 0mm 0mm 0mm 0mm, clip, width=2.2cm]{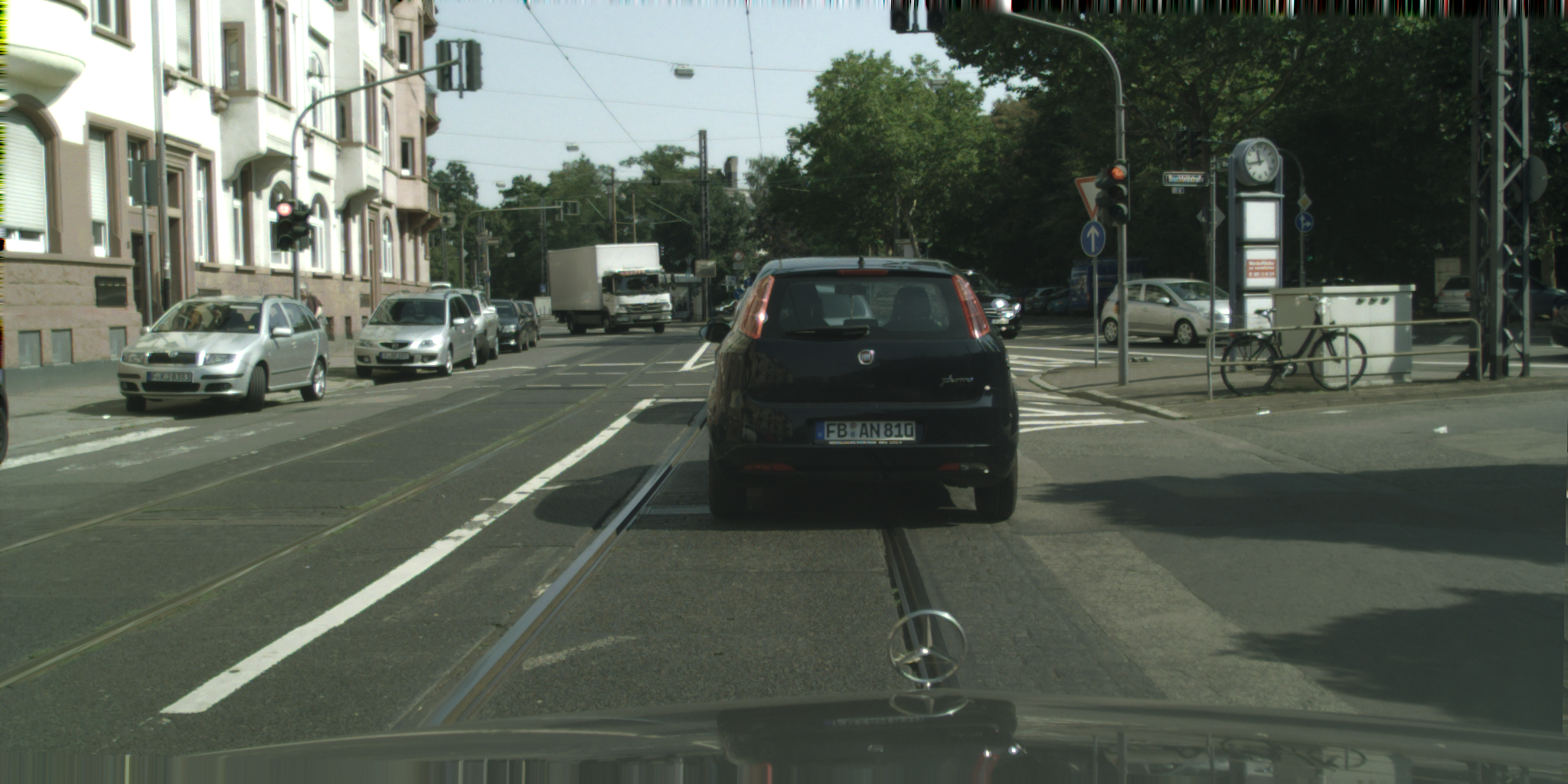} &
  \includegraphics[trim = 0mm 0mm 0mm 0mm, clip, width=2.2cm]{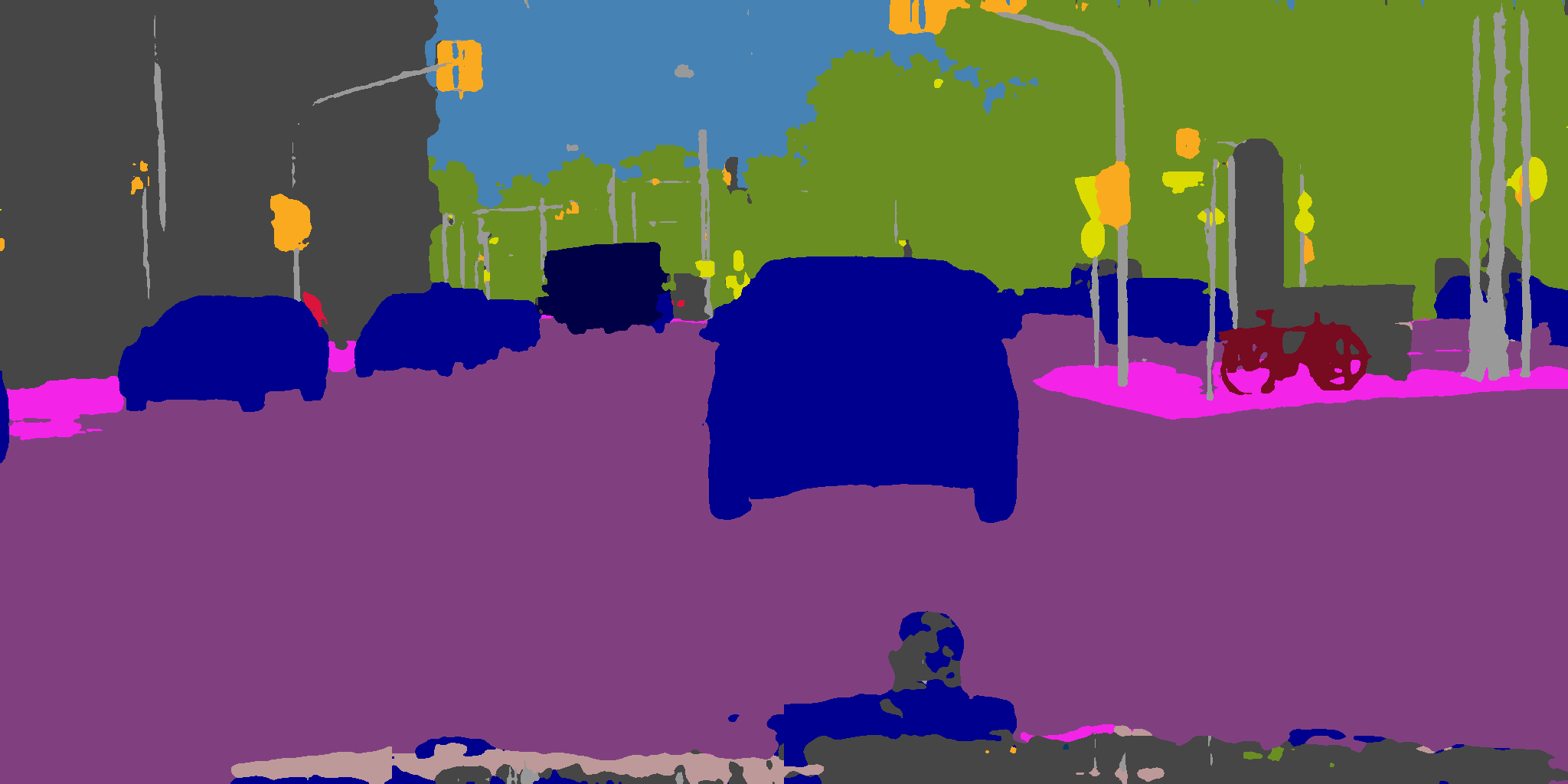} &
  \includegraphics[trim = 0mm 0mm 0mm 0mm, clip, width=2.2cm]{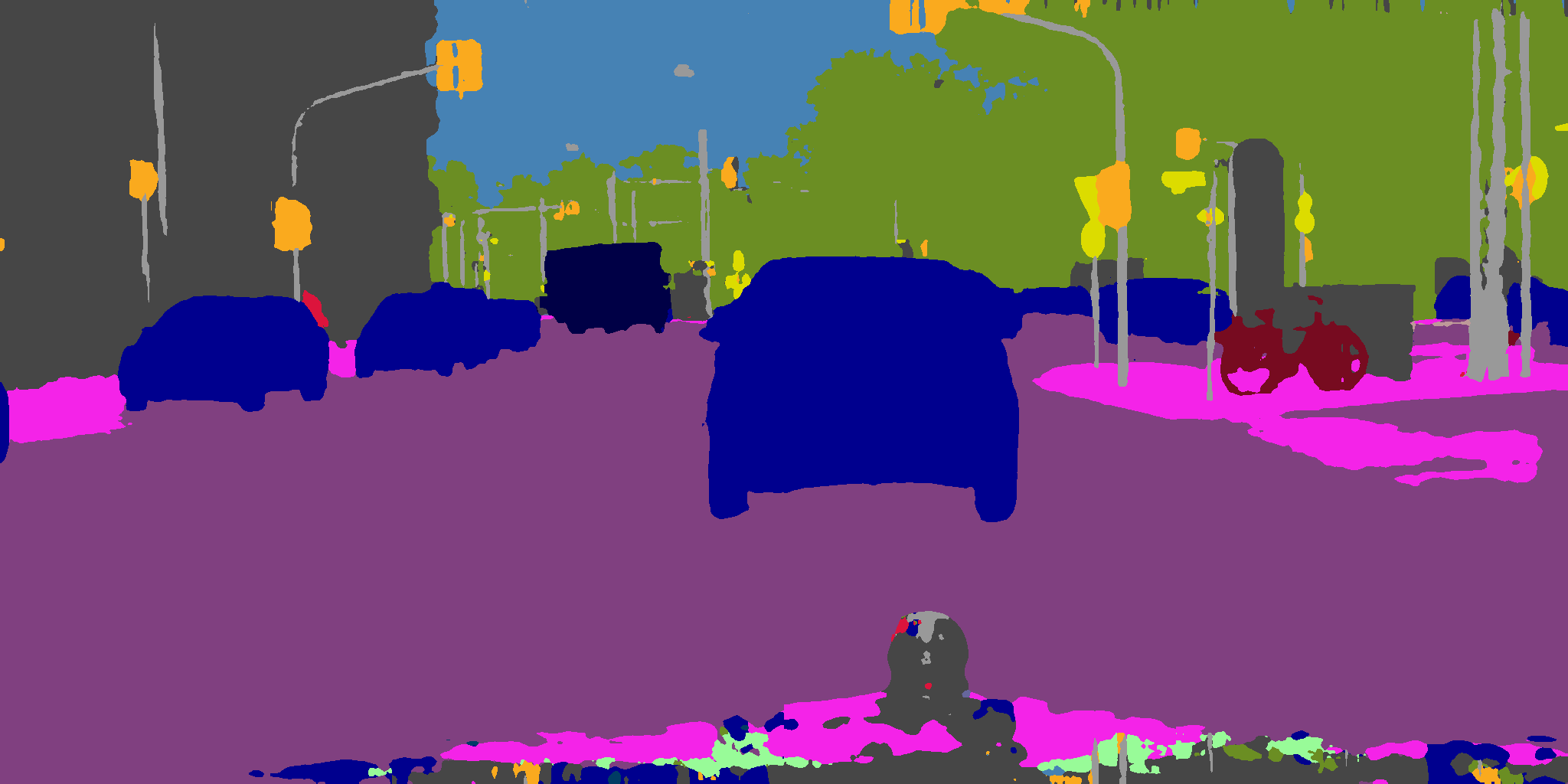} &
  \includegraphics[trim = 0mm 0mm 0mm 0mm, clip, width=2.2cm]{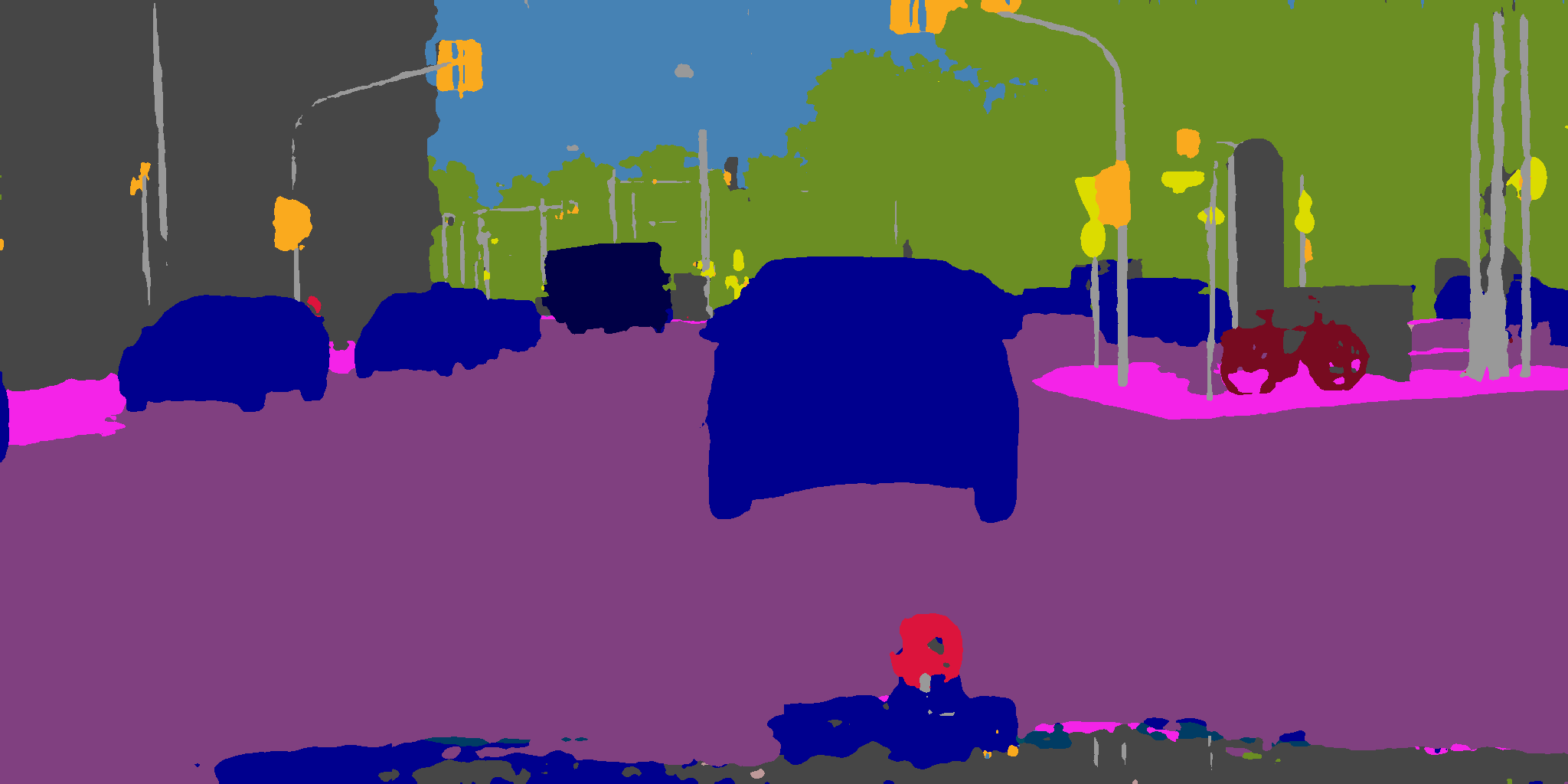} &
  \includegraphics[trim = 0mm 0mm 0mm 0mm, clip, width=2.2cm]{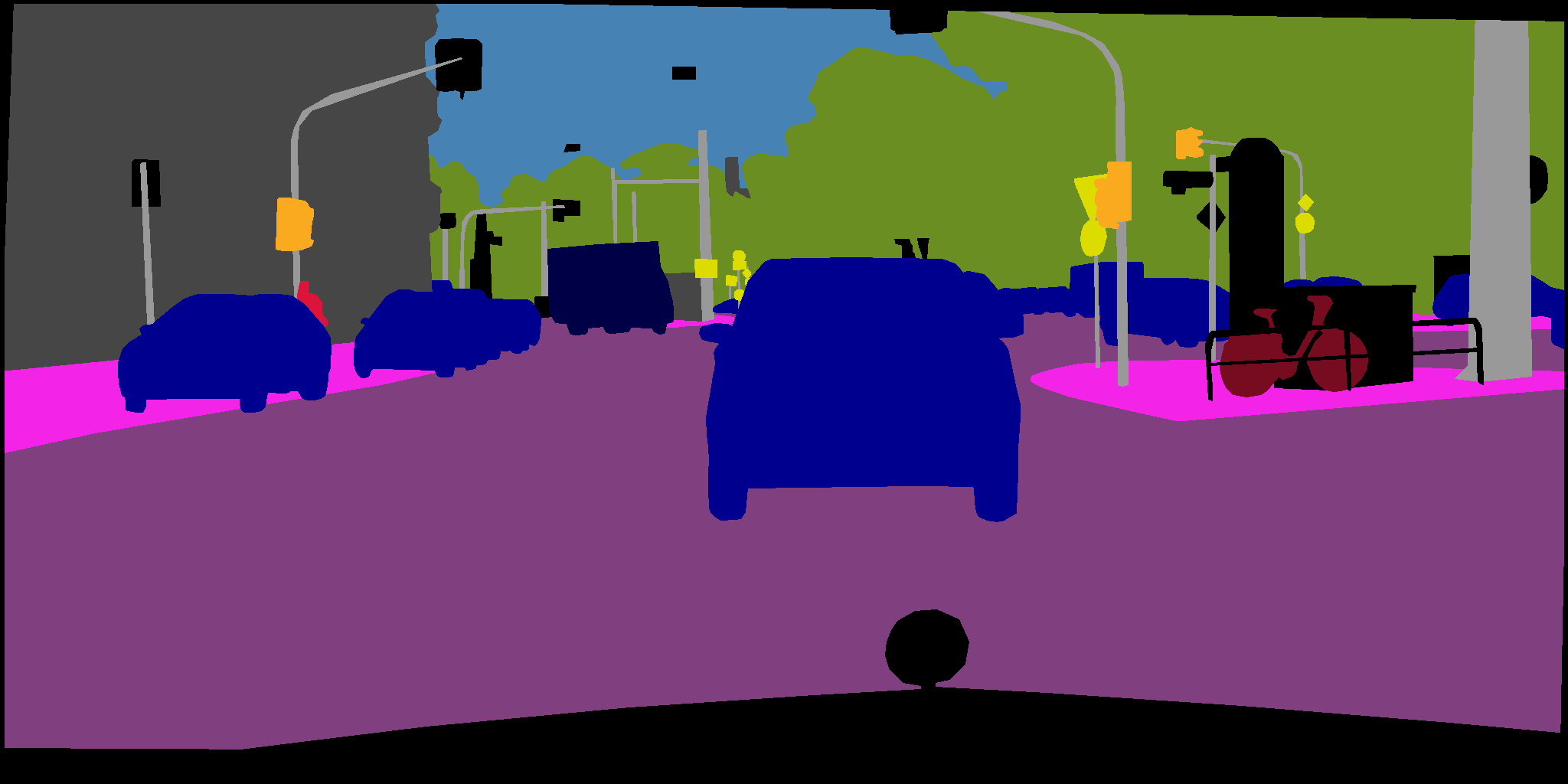} \\
  
  \includegraphics[trim = 0mm 0mm 0mm 0mm, clip, width=2.2cm]{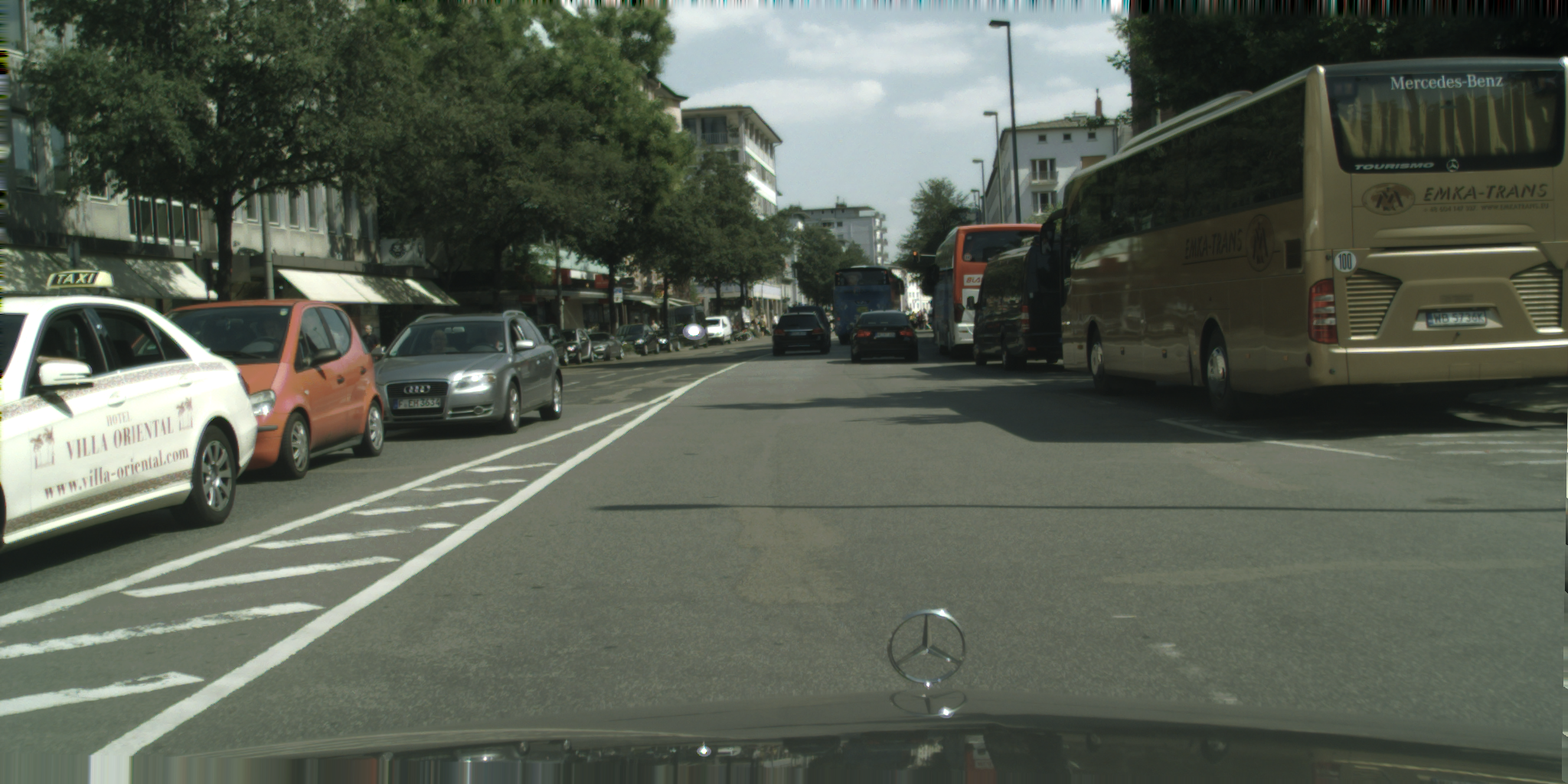} &
  \includegraphics[trim = 0mm 0mm 0mm 0mm, clip, width=2.2cm]{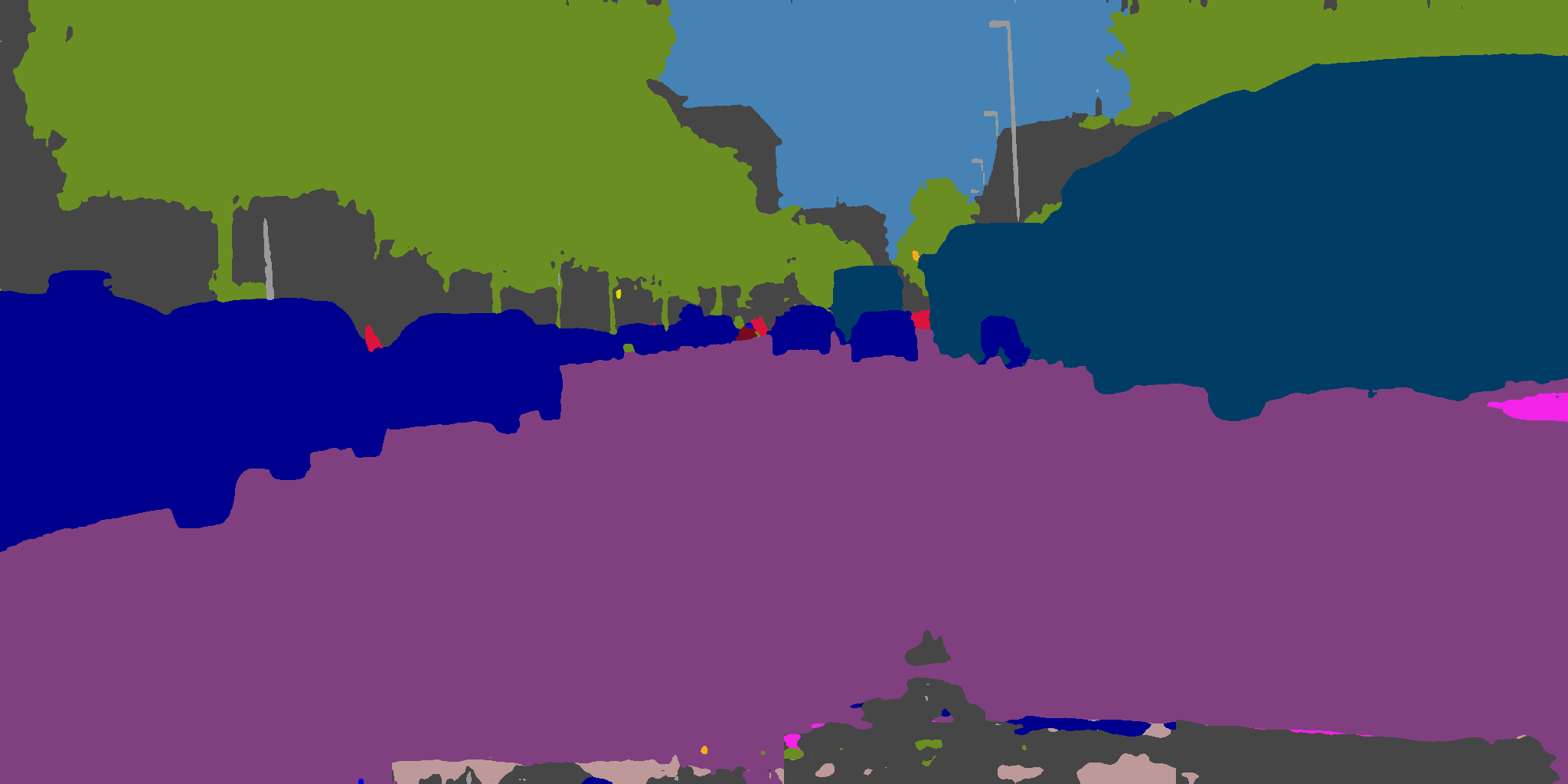} &
  \includegraphics[trim = 0mm 0mm 0mm 0mm, clip, width=2.2cm]{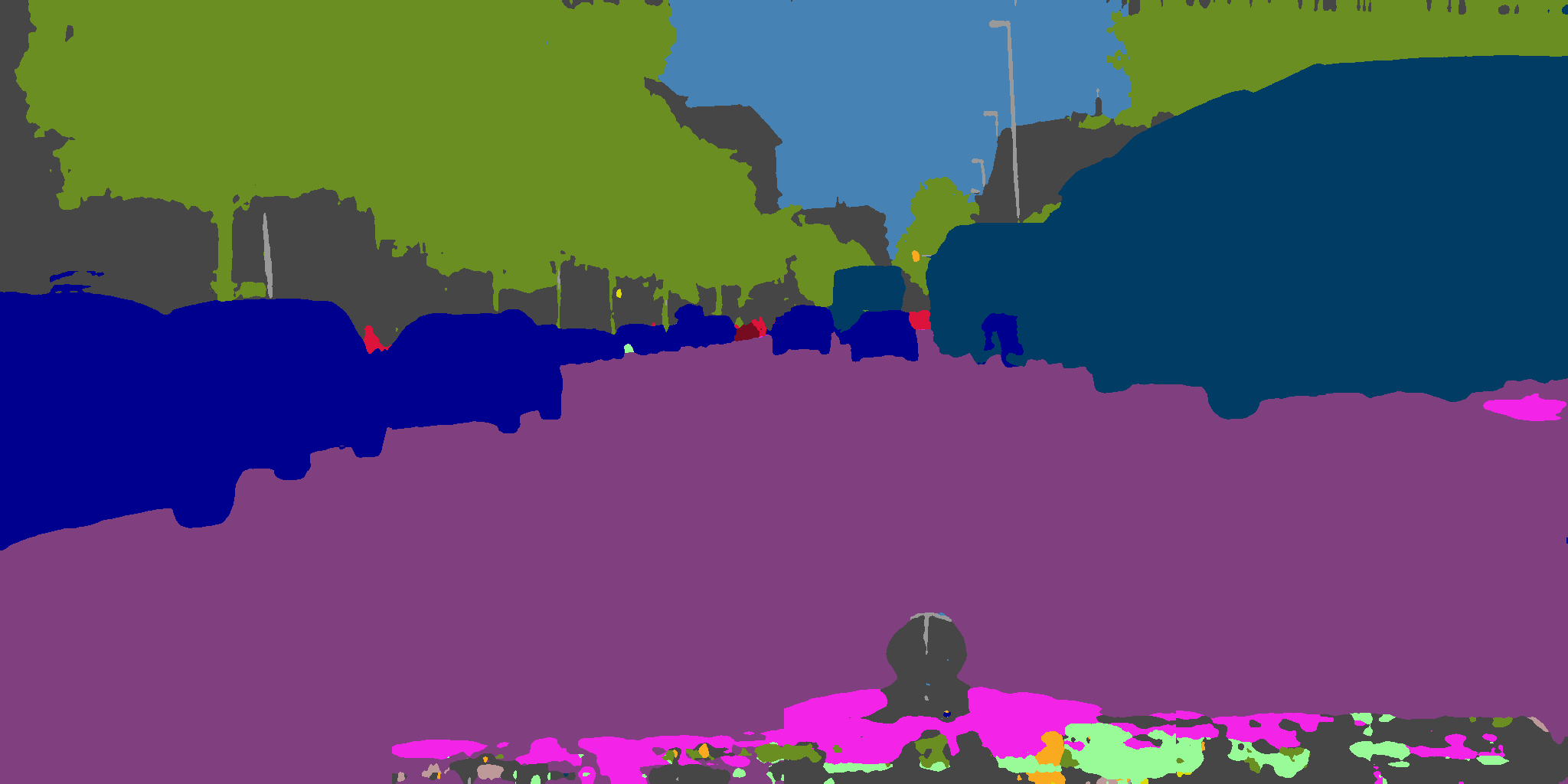} &
  \includegraphics[trim = 0mm 0mm 0mm 0mm, clip, width=2.2cm]{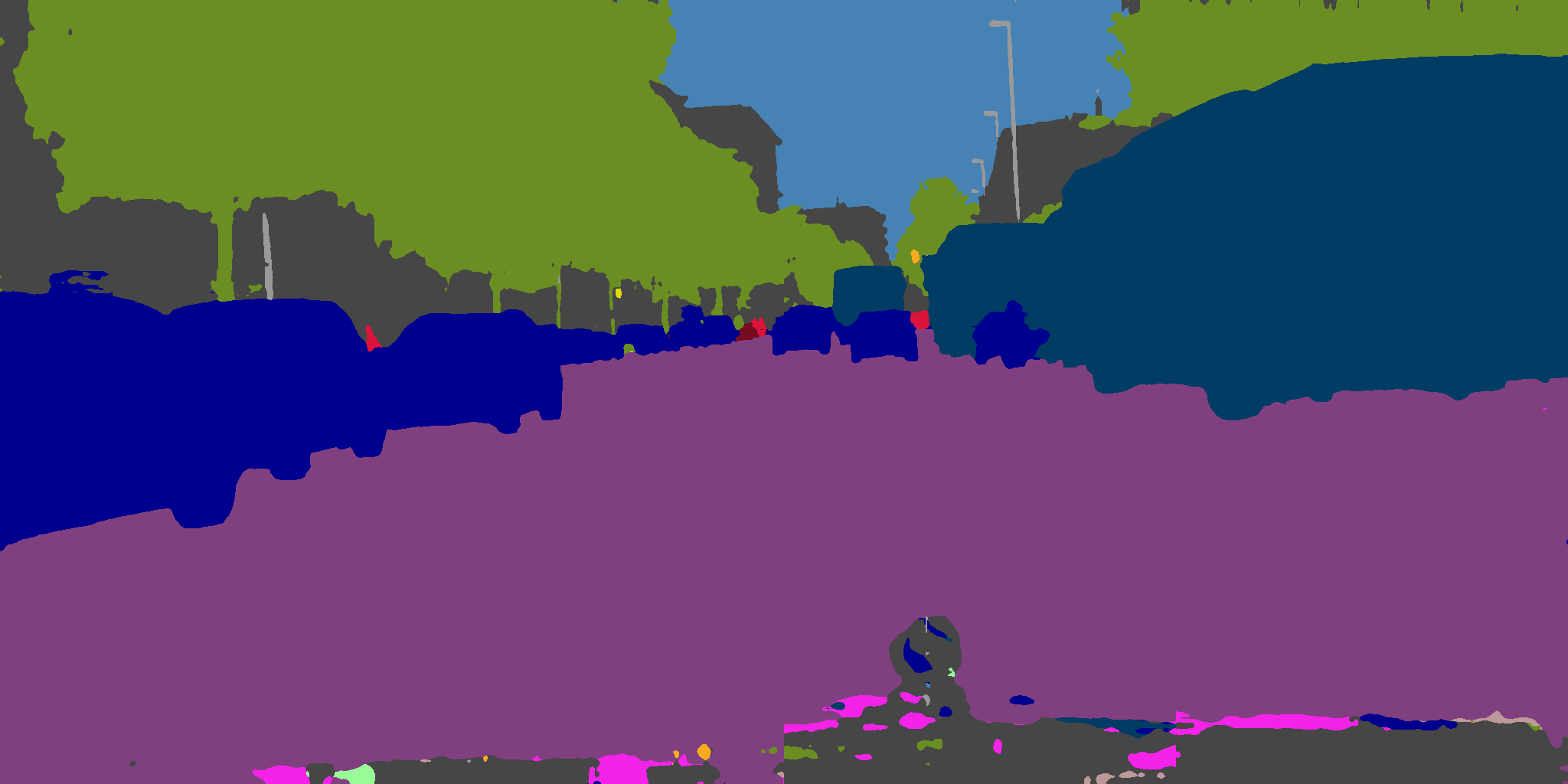} & 
  \includegraphics[trim = 0mm 0mm 0mm 0mm, clip, width=2.2cm]{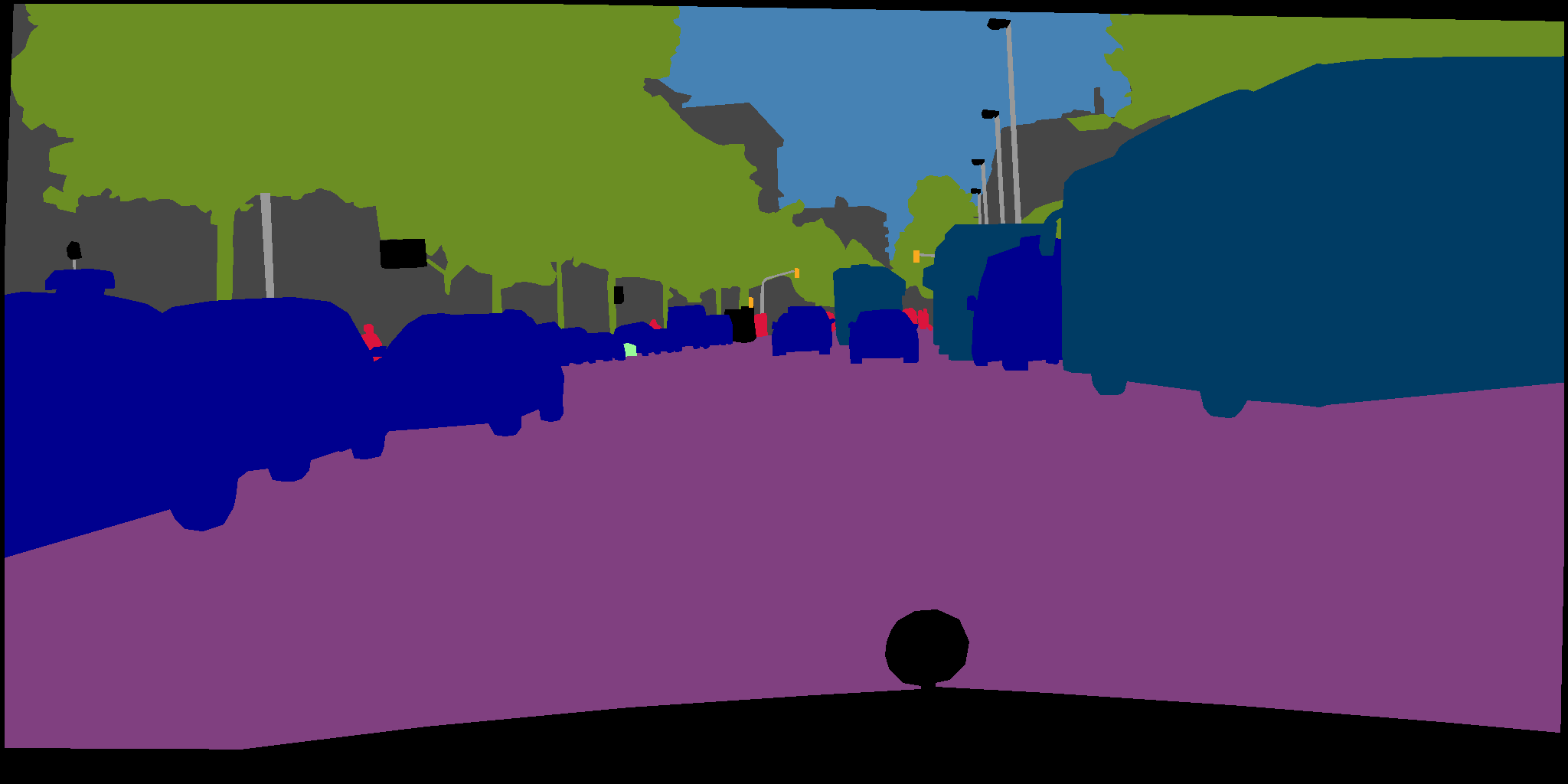} \\
  
  \includegraphics[trim = 0mm 0mm 0mm 0mm, clip, width=2.2cm]{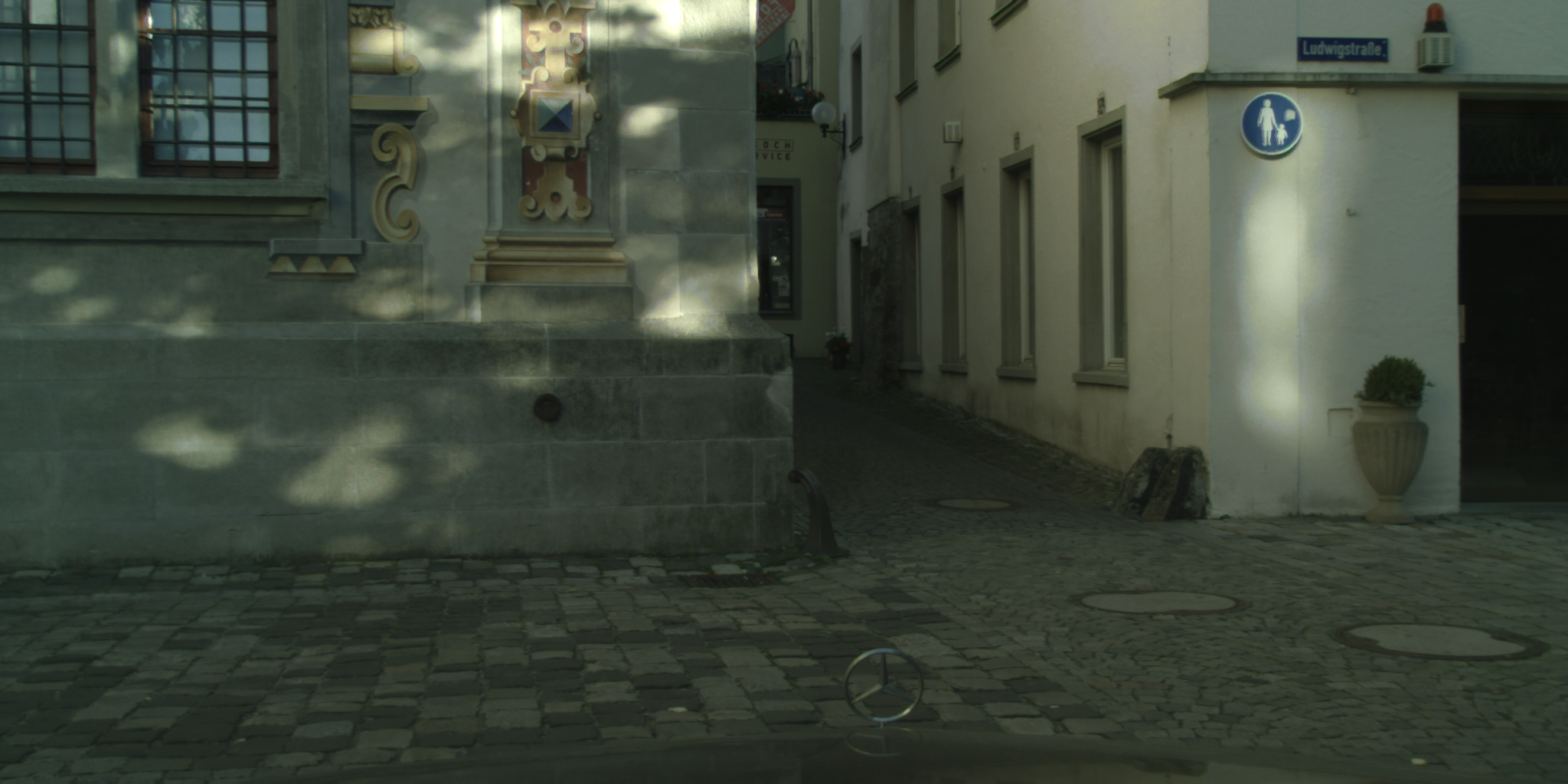} &
  \includegraphics[trim = 0mm 0mm 0mm 0mm, clip, width=2.2cm]{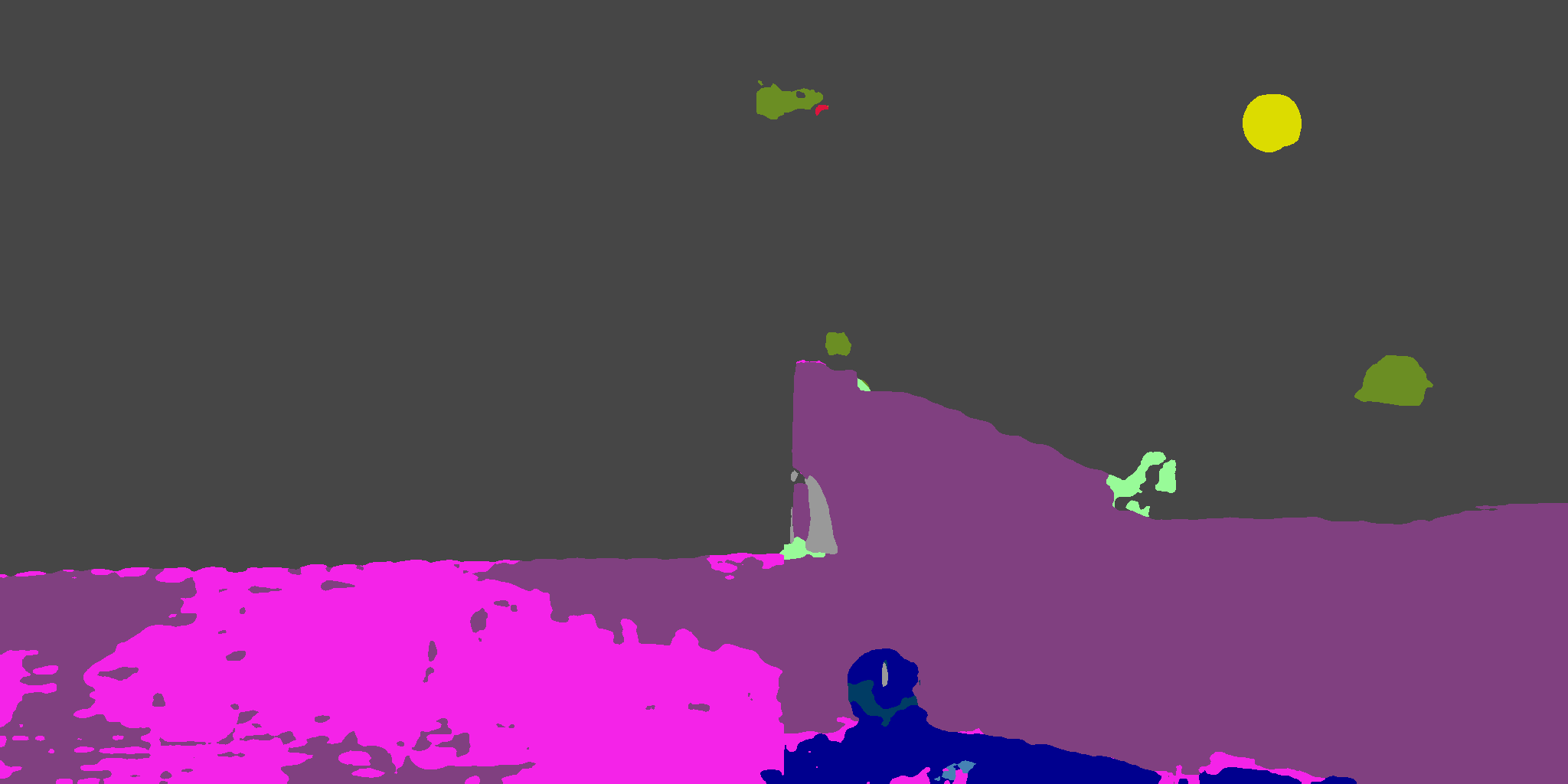} &
  \includegraphics[trim = 0mm 0mm 0mm 0mm, clip, width=2.2cm]{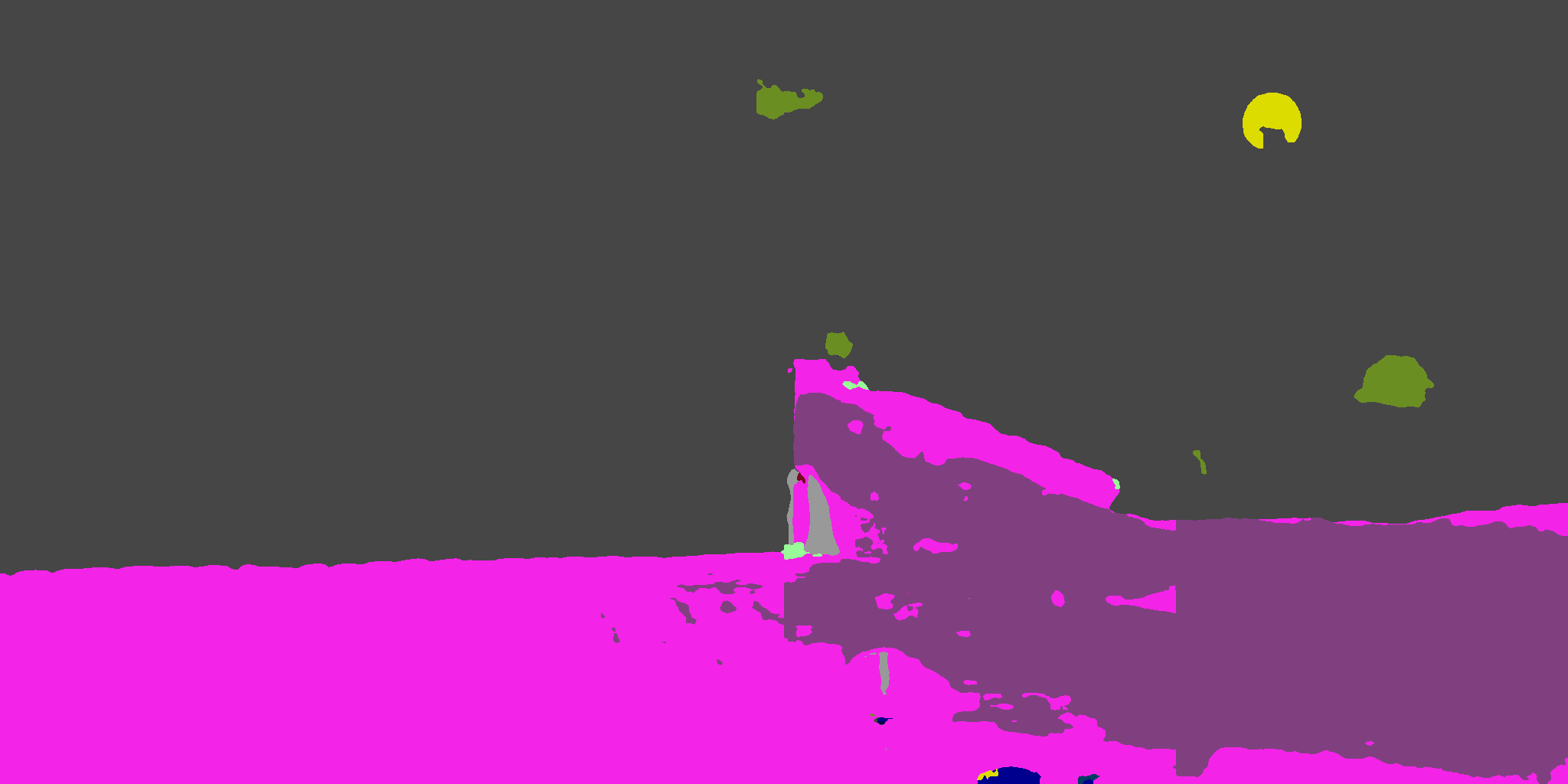} &
  \includegraphics[trim = 0mm 0mm 0mm 0mm, clip, width=2.2cm]{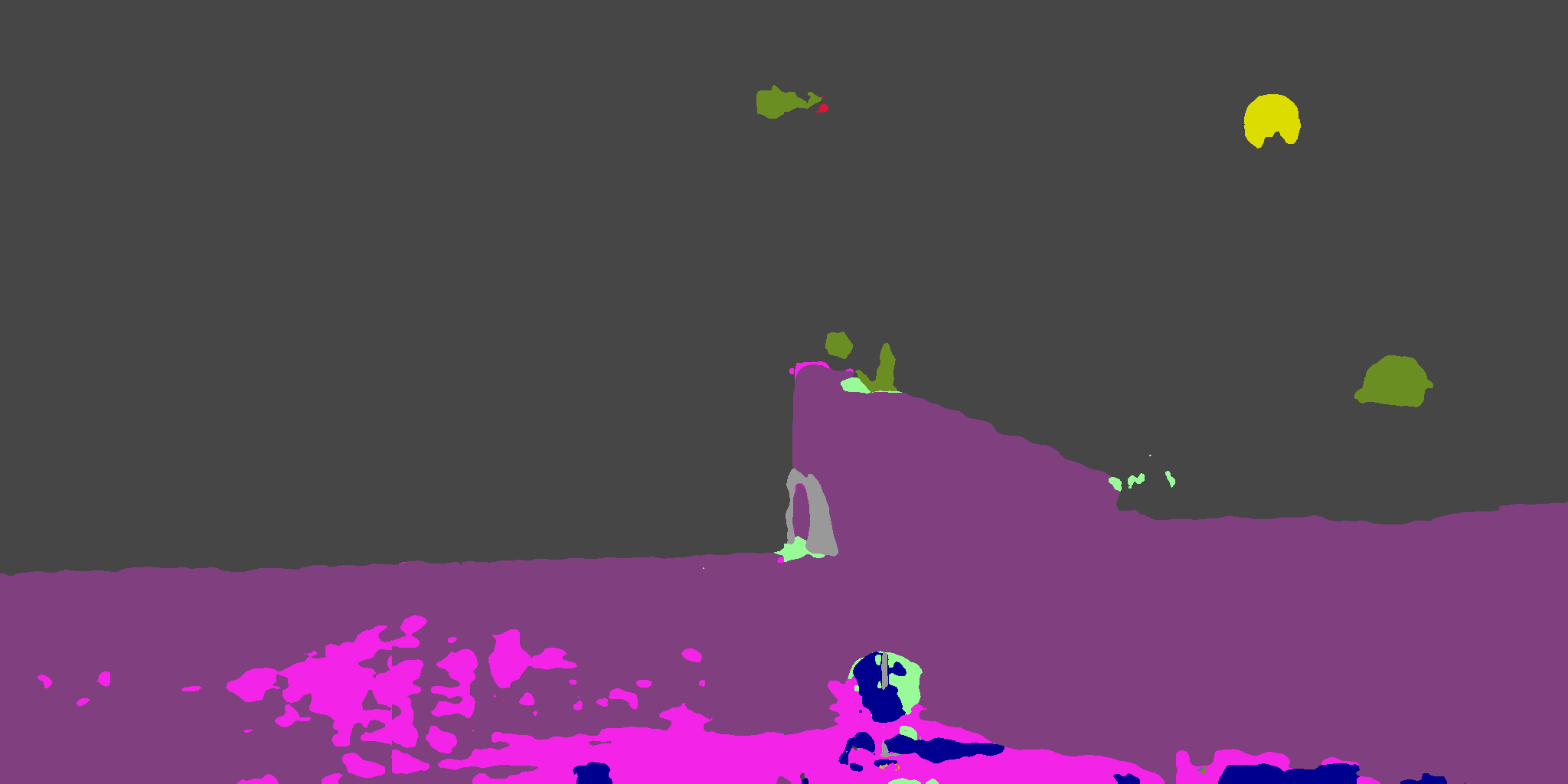} &
  \includegraphics[trim = 0mm 0mm 0mm 0mm, clip, width=2.2cm]{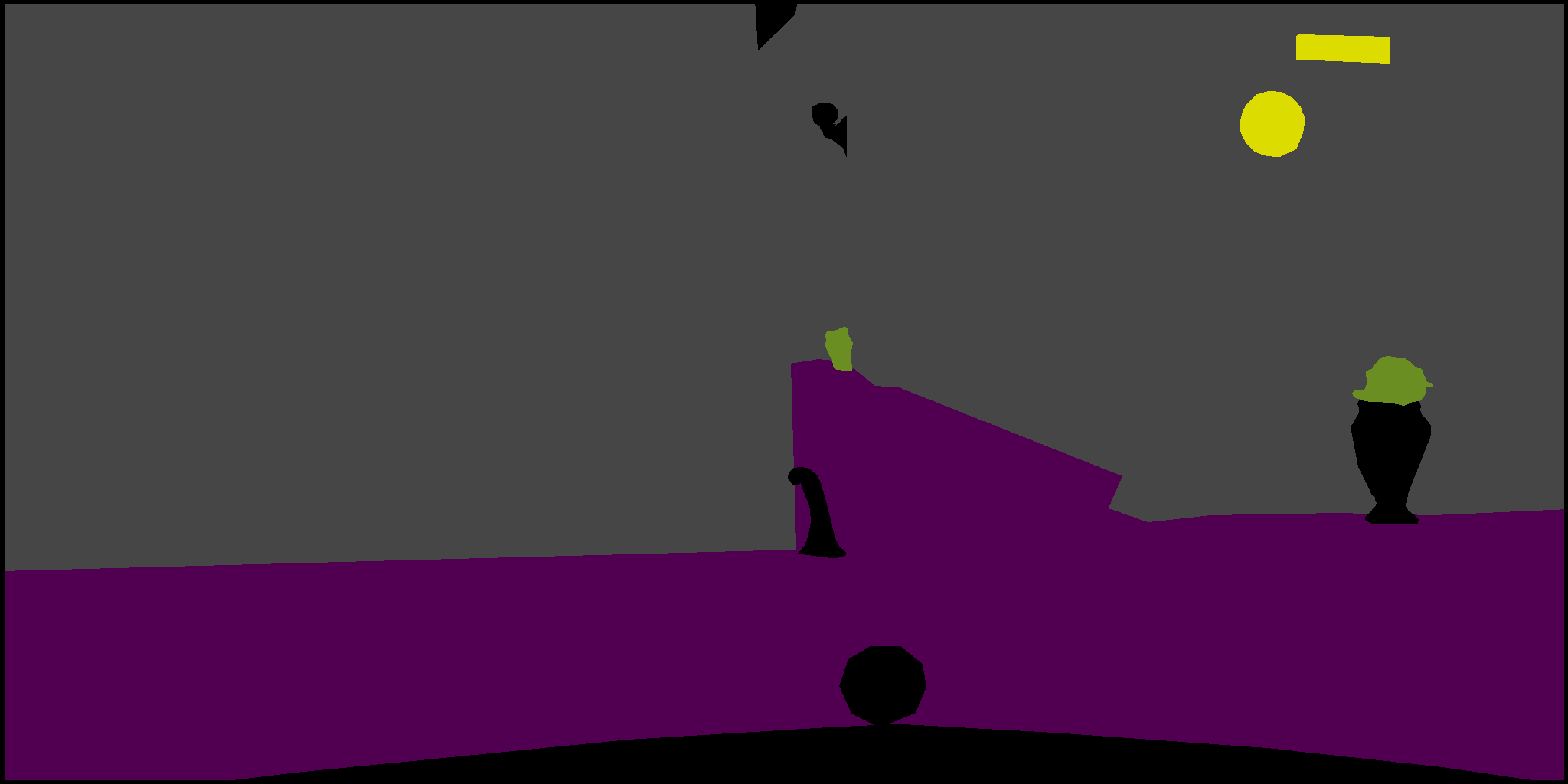} \\
  
\end{tabular}}
\caption{
Segmentation predictions on GTA$\rightarrow$CS validation set. Each column from left to right shows the Cityscapes image, HRDA \cite{Hoyer22eccv} prediction, MIC \cite{Hoyer23} prediction, CoPT (our) prediction and the ground truth label. In all examples, CoPT produces fewer spurious predictions in the sidewalk, showing its benefits for classes with large intra-class variation.}
\label{fig:segmentations_cityscapes}
\end{figure*}

 We compare CoPT to previous UDA methods for semantic segmentation on GTA$\rightarrow$CS in Table \ref{tab:gta2cs}. The methods are trained using the GTA training set as the source domain and evaluated in terms of IOU on the Cityscapes validation set. CoPT gives a 1.6 \% boost in mIOU when trained on top of MIC, and shows improvements on most classes, setting a new state of the art on the benchmark. CoPT gives the highest improvement on the train class: the train's large size means it is made up of a larger variety of features than other classes, so it might benefit more from associating the class with the centroid of a wide variety of semantic features. Figure \ref{fig:segmentations_cityscapes} shows qualitative results on Cityscapes from CoPT and two other baselines. It illustrates how CoPT can improve performance on the sidewalk class, which also exhibits high intra-class variation. CoPT struggles most with the traffic sign class: this class has more varied geometric shapes than the other classes and its accurate segmentation depends more on accurate boundary localization rather than on distinguishing domain textures.

\par To test how CoPT compares to previous methods on a different source dataset, we train it using Synthia synthetic data and test it on the Cityscapes validation set, reporting IOU for all classes in Table \ref{tab:syn2cs}. Synthia data constitutes a larger domain difference from Cityscapes than GTA, largely due to the camera angles being from above rather than from a driver's perspective. Regardless of this challenge, CoPT is able to surpass MIC's performance as well as previous methods' performance on most classes. CoPT performs particularly well on the motorbike/motorcycle class. However, it struggles most with the sidewalk and road classes. In many Synthia source images, foliage is placed in front of the camera and blocks the sidewalk and roads. Since CoPT relies on downsampled ground truth masks to extract pixel features for a class, when there are thin structures in front of large classes this can lead to mixing class features in the extracted pixel feature, explaining CoPT's struggle on sidewalks and roads. 
\par To show the generalizability of CoPT, we evaluate its performance on CS$\rightarrow$DZ, a benchmark that considers adaptation from daytime images to nighttime. CoPT outperforms previous methods on overall mIOU, and consistently attains top one to two performance on nearly all categories. It achieves the best performance on the truck class and also performs particularly well on the sidewalk and terrain classes compared to previous methods. This affirms our findings on the GTA$\rightarrow$CS benchmark in Table \ref{tab:gta2cs} that CoPT is improving intra-class understanding, since these classes tend to be large in the image, containing numerous visual features while making up the same class.
\par To test whether CoPT performs well on another target domain we train it with labeled Cityscapes training images and unlabeled ACDC training images and test it on the adverse conditions in the ACDC test set as the target domain. Results on each condition are presented in Table \ref{tab:cs2acdc} where CoPT is compared to MIC. CoPT outperforms MIC on average due to its 3.6\% improvement on the Fog condition and maintains similar performance on other conditions. Compared to the other adverse conditions, fog tends to uniformly change class appearance, whereas a condition like nighttime can drastically change appearance with lights illuminating different object parts and obscuring others, which supports the hypothesis that CoPT improves intra-class understanding.

\begin{table}[t]
\begin{minipage}{\textwidth}
  \caption{\scriptsize{
  Semantic segmentation results reported as \% IOU on the GTA$\rightarrow$CS val set. \textbf{Bold} indicates highest performance and \underline{underline} indicates second highest. All methods' results are taken from the published paper except for those labeled with $\dagger$, which indicates the method was retrained. Our method, built on top of MIC \cite{Hoyer23}, attains state of the art performance.
  }}
  \label{tab:gta2cs}
  \vspace{-1em}
  \centering
  \resizebox{\textwidth}{!}{%
  \begin{tabular}{@{}lccccccccccccccccccc|c@{}}
    \toprule
    Method & Road & S.Walk & Build. & Wall & Fence & Pole & T.Light & T.Sign & Veg. & Terr. & Sky & Person & Rider & Car & Truck & Bus & Train & M.Bike & Bike & \textbf{mIOU} \\
    \midrule
    DAFormer \cite{Hoyer22cvpr} & 95.7 & 70.2 & 89.4 & 53.5 & 48.1 & 49.6 & 55.8 & 59.4 & 89.9 & 47.9 & 92.5 & 72.2 & 44.7 & 92.3 & 74.5 & 78.2 & 65.1 & 55.9 & 61.8 & 68.3 \\
    SePiCo \cite{Xie23} & 95.2 & 67.8 & 88.7 & 41.4 & 38.4 & 43.4 & 55.5 & 63.2 & 88.6 & 46.4 & 88.3 & 73.1 & 49.0 & 91.4 & 63.2 & 60.4 & 0.0 & 45.2 & 60.0 & 61.0 \\
    HRDA \cite{Hoyer22eccv}  & 96.4 & 74.4 & 91.0 & 61.6 & 51.5 & 57.1 & 63.9 & 69.3 & 91.3 & 48.4 & 94.2 & 79.0 & 52.9 & 93.9 & 84.1 & 85.7 & 75.9 & 63.9 & 67.5 & 73.8 \\
    PiPa (HRDA) \cite{Chen23pipa} & 96.8 & 76.3 & \textbf{91.6} & \textbf{63.0} & \textbf{57.7} & \textbf{60.0} & \underline{65.4} & \underline{72.6} & \underline{91.7} & \underline{51.8} & \textbf{94.8} & \underline{79.7} & \textbf{56.4} & 94.4 & 85.9 & \underline{88.4} & \underline{78.9} & 63.5 & 67.2 & \underline{75.6} \\
    $\text{MIC}^{\dagger}$ \cite{Hoyer23} & \underline{97.5} & \underline{80.3} & \underline{91.3} & 60.2 & 52.5 & \underline{59.7} & 64.1 & \textbf{73.1} & 91.3 & 49.6 & 93.9 & 79.3 & 54.8 & \underline{94.6} & \underline{86.5} & 86.6 & 67.2 & \textbf{66.8} & \underline{68.7} & 74.6 \\
    \midrule
    $\textbf{CoPT (Ours)}$ & \textbf{97.6} & \textbf{80.9} & \textbf{91.6} & \underline{62.1} & \underline{55.9} & 59.3 & \textbf{66.7} & 70.5 & \textbf{91.9} & \textbf{53.0} & \underline{94.4} & \textbf{80.0} & \underline{55.6} & \textbf{94.7} & \textbf{87.1} & \textbf{88.6} & \textbf{82.1} & \underline{65.0} & \textbf{68.8} & \textbf{76.1} \\
  \bottomrule
  \end{tabular}}
  \vspace{0.5em}
\end{minipage}  
\begin{minipage}{\textwidth}
  \caption{\scriptsize{
  Semantic segmentation results reported as \% IOU on the Synthia$\rightarrow$CS val set. All methods' results are taken from the published paper except for those labeled with $\dagger$, which indicates the method was retrained. Our method attains state of the art performance.
  }}
  \label{tab:syn2cs}
  \vspace{-1em}
  \centering
  \resizebox{\textwidth}{!}{%
  \begin{tabular}{@{}lcccccccccccccccc|c@{}}
    \toprule
    Method & Road & S.Walk & Build. & Wall & Fence & Pole & T.Light & T.Sign & Veg. & Sky & Person & Rider & Car & Bus & M.Bike & Bike & \textbf{mIOU} \\
    \midrule
    DAFormer \cite{Hoyer22cvpr} & 84.5 & 40.7 & 88.4 & 41.5 & 6.5 & 50.0 & 55.0 & 54.6 & 86.0 & 89.9 & 73.2 & 48.2 & 87.2 & 53.2 & 53.9 & 61.7 & 60.9 \\
    SePiCo \cite{Xie23} & 77.0 & 35.3 & 85.1 & 23.9 & 3.4 & 38.0 & 51.0 & 55.1 & 85.6 & 80.5 & 73.5 & 46.3 & 87.6 & \textbf{69.7} & 50.9 & \textbf{66.5} & 58.1 \\
    HRDA \cite{Hoyer22eccv} & \underline{85.2} & 47.7 & 88.8 & \underline{49.5} & 4.8 & 57.2 & 65.7 & 60.9 & 85.3 & 92.9 & 79.4 & 52.8 & 89.0 & 64.7 & 63.9 & 64.9 & 65.8 \\
    $\text{PiPa (HRDA)}^{\dagger}$ \cite{Chen23pipa} & \textbf{86.6} & \textbf{52.7} & 88.2 & 46.6 & 1.4 & 47.8 & 60.3 & 46.2 & 86.7 & 93.2 & 75.5 & 50.8 & 86.6 & 17.0 & 59.8 & 63.6 & 67.3 \\
    MIC \cite{Hoyer23} & \textbf{86.6} & \underline{50.5} & \underline{89.3} & 47.9 & \underline{7.8} & \underline{59.4} & \underline{66.7} & \textbf{63.4} & \underline{87.1} & \underline{94.6} & \underline{81.0} & \textbf{58.9} & \textbf{90.1} & 61.9 & \underline{67.1} & 64.3 & \underline{67.3} \\
    \midrule
    $\textbf{CoPT (Ours)}$ & 83.4 & 44.3 & \textbf{90.0} & \textbf{50.4} & \textbf{8.0} & \textbf{60.0} & \textbf{67.0} & \underline{63.0} & \textbf{87.5} & \textbf{94.8} & \textbf{81.1} & \underline{58.6} & \underline{89.7} & \underline{66.5} & \textbf{68.9} & \underline{65.0} & \textbf{67.4} \\
  \bottomrule
  \end{tabular}}
  \vspace{0.5em}
\end{minipage}  
\begin{minipage}{\textwidth}
  \caption{\scriptsize{
  Semantic segmentation results reported as \% IOU on the CS$\rightarrow$DZ test set. \textbf{Bold} indicates highest performance and \underline{underline} indicates second highest. All methods' results are taken from the publication except for those labeled with $\dagger$, which indicates the method was retrained. 
  }}
  \label{tab:cs2dz}
  \vspace{-1em}
  \centering
  \resizebox{\textwidth}{!}{%
  \begin{tabular}{@{}lccccccccccccccccccc|c@{}}
    \toprule
    Method & Road & S.Walk & Build. & Wall & Fence & Pole & T.Light & T.Sign & Veg. & Terr. & Sky & Person & Rider & Car & Truck & Bus & Train & M.Bike & Bike & \textbf{mIOU} \\
    \midrule
    DAFormer \cite{Hoyer22cvpr} & \textbf{93.5} & \underline{65.5} & 73.3 & 39.4 & 19.2 & 53.3 & 44.1 & 44.0 & \underline{59.5} & 34.5 & 66.6 & 53.4 & 52.7 & \underline{82.1} & 52.7 & \underline{9.5} & 89.3 & \textbf{50.5} & 38.5 & 53.8 \\
    SePiCo \cite{Xie23} & 91.2 & 61.3 & 67.0 & 28.5 & 15.5 & 44.7 & 44.3 & 41.3 & \textbf{65.4} & 22.5 & \textbf{80.4} & 41.3 & 52.4 & 71.2 & 39.3 & 0.0 & 39.6 & 27.5 & 28.8 & 45.4 \\
    HRDA \cite{Hoyer22eccv} & 90.4 & 56.3 & 72.0 & 39.5 & \underline{19.5} & 57.8 & \underline{52.7} & 43.1 & 59.3 & 29.1 & \underline{70.5} & 60.0 & \textbf{58.6} & \textbf{84.0} & \underline{75.5} & \textbf{11.2} & 90.5 & 51.6 & 40.9 & \underline{55.9} \\
    $\text{MIC}^{\dagger}$ \cite{Hoyer23} & 89.1 & 54.0 & \textbf{80.4} & \textbf{53.5} & \textbf{20.3} & \textbf{63.8} & \textbf{54.9} & \textbf{54.2} & 47.9 & \underline{37.9} & 47.6 & \textbf{63.9} & \underline{58.3} & 75.2 & 66.5 & 7.3 & \underline{90.7} & 48.9 & \textbf{45.9} & 55.8 \\
    $\textbf{CoPT (Ours)}$ & \underline{92.7} & \textbf{66.1} & \underline{80.0} & \underline{49.0} & 19.3 & \underline{63.2} & 51.7 & \underline{52.0} & 50.9 & \textbf{43.1} & 55.9 & \underline{61.7} & 56.5 & 59.6 & \textbf{79.4} & 2.96 & \textbf{90.8} & \underline{50.0} & \underline{42.5} & \textbf{56.2} \\
  \bottomrule
  \end{tabular}}
\end{minipage}
\vspace{-0.5em}
\end{table}

\begin{table}[t]
  \caption{\scriptsize
  Semantic segmentation results reported as \% IOU on the adverse conditions in CS$\rightarrow$ACDC test set. $\dagger$ indicates MIC \cite{Hoyer23} was retrained. CoPT outperforms MIC on average and especially on fog.
  }
  \label{tab:cs2acdc}
  \vspace{-1.0em}
  \centering
  \small
  \def\arraystretch{0.9}  
  \setlength\tabcolsep{0.6em}  
  \scalebox{0.75}{
  \begin{tabular}{@{}lcccc|c@{}}
    \toprule
    Method & Rain & Snow & Fog & Night & \textbf{All} \\
    \midrule
    $\text{MIC}^{\dagger}$ \cite{Hoyer23} & \textbf{74.2} & \textbf{60.7} & 56.8 & \textbf{57.6} & 63.3 \\
    \midrule
    $\textbf{CoPT (Ours)}$ & 74.0 & 60.6 & \textbf{60.4} & 56.8 & \textbf{63.7} \\
  \bottomrule
  \end{tabular}}
\end{table}

\subsection{Ablations}
\label{sec:ablations}

\begin{table}[t]
\begin{minipage}{\textwidth}
\begin{minipage}{0.47\textwidth}
  \caption{Ablation experiment results for CoPT's covariance distance metric are reported as \% mIOU on GTA$\rightarrow$CS over 19 classes.
  }
  \label{tab:copt_metric_ablation}
  \centering
  \vspace{-1.0em}
  \begin{tabular}{@{}c@{\hskip 0.5cm}c@{}}
    \toprule
    Metric & mIOU \\
    \midrule
    L1 & 75.34 \\
    L2 & 70.18 \\
    Cosine Similarity & \textbf{76.08} \\
  \bottomrule
  \end{tabular}
\end{minipage}
\hspace{0.04\textwidth}
\begin{minipage}{0.47\textwidth}
    \caption{Ablation experiment results for LLM Domain Templates vs. hand -crafted templates are reported as \% mIOU on GTA$\rightarrow$CS over 19 classes.
  }
  \label{tab:template_ablation}
  \centering
  \begin{tabular}{@{}c@{\hskip 0.5cm}c@{}}
    \toprule
    Template Type & mIOU \\
    \midrule
    Hand-Crafted & 75.95 \\
    LLM Generation & \textbf{76.08} \\
  \bottomrule
  \end{tabular}  
\end{minipage}
\end{minipage}
%
%

%
%

%

%
\begin{minipage}{0.47\textwidth}
  \caption{Ablation experiment results for CoPT added to two different UDA methods. Results reported as \% mIOU on GTA$\rightarrow$CS over 19 classes.
  }
  \label{tab:copt_backbone_ablation}
  \vspace{-1.0em}
  \centering
  \scalebox{0.95}{
  \begin{tabular}{@{}c@{\hskip0.4cm}c@{\hskip0.5cm}c@{}}
    \toprule
    Method & mIOU & $\Delta$ \\
    \midrule
    HRDA & 73.8 & - \\
    HRDA+CoPT & \textbf{75.34} & +1.54 \\
    \midrule
    MIC & 74.61 & - \\
    MIC+CoPT & \textbf{76.08} & +1.47 \\
  \bottomrule
  \end{tabular}
  }
\end{minipage}
\hspace{0.04\textwidth}
\begin{minipage}{0.47\textwidth}
\caption{CoPT results in \% mIOU on GTA$\rightarrow$CS val 19 classes using different text encoders.
}
\label{tab:text_encoder}
\centering
\small
\def\arraystretch{1.0}  
\setlength\tabcolsep{0.6em}  
\scalebox{1}{
\begin{tabular}{cc}
\toprule
Text Encoder & mIOU \\
\midrule
\textcolor{gray}{MIC Baseline} & \textcolor{gray}{74.6} \\
Mistral & 70.8 \\
Sentence-T5 & 75.5 \\
CLIP & \textbf{76.1} \\
\bottomrule
\end{tabular}}
\end{minipage}
%
%

\begin{minipage}{\textwidth}
\caption{CoPT mIOU (\%) across 19 classes on GTA$\rightarrow$CS val set using pixel features extracted from source, pseudo-labeled target, or both domains.
}
\label{tab:copt_target}
\centering
\small
\def\arraystretch{0.9}  
\setlength\tabcolsep{0.6em}  
\scalebox{1}{
\begin{tabular}{cc}
\toprule
CoPT Pixel Features & mIOU \\ \midrule
\textcolor{gray}{MIC Baseline} & \textcolor{gray}{74.6} \\
Source + Target & 71.8 \\
Target Only & 74.7 \\
Source Only & \textbf{76.1} \\ \bottomrule
\end{tabular}
}
\end{minipage}
\end{table}
\par \textbf{CoPT Distance Metric} We compare different metrics to calculate the distance between CoPT's pixel and text covariance matrices in Table \ref{tab:copt_metric_ablation}. Using a cosine similarity loss outperforms both L1 and L2 distances. L2 considers the covariance between each pair of classes equally, which may not be ideal since certain class pairs can co-occur more often in target vs. source domains and thus contribute more to the domain gap. L1 is ideal for scenarios involving grid-like space, which is not similar to the covariance matrices since different batches contain different classes. Cosine similarity favors points that share the same direction, which is conducive to our goal of aligning representations in abstract latent spaces with unknown structure.
%
\par \noindent 
\textbf{LLM Domain Template vs. Hand-Crafted Templates} Table \ref{tab:template_ablation} shows our evaluation of whether LLM Domain Template, in which an LLM generates domain descriptions, leads to better domain-agnostic features for CoPT than simple hand-crafted domain descriptions. The evaluation is done on the GTA$\rightarrow$CS benchmark. While it shows that LLM Domain Template outperforms the hand-crafted approach, the slight improvement could indicate that the average hand-crafted template embedding is not very far from the average LLM Domain Template embedding. Averaging templates gives the centroid of the points, and since the domain-specific hand-crafted templates have so far been designed to be quite general already, their centroid will likely also be domain-agnostic. However, this can pose a problem in cases where access to the target domain is not allowed, in which case it is beneficial to have our automated method generate descriptions of the target. This can be accomplished by feeding a target image in addition to the text query in Figure \ref{fig:text_embedding_generation} in LLM Domain Template.
\par \noindent \textbf{Backbone Method} We test whether CoPT is able to boost performance of two different UDA methods, MIC and HRDA, for semantic segmentation by jointly training these methods with our loss. The results on the GTA$\rightarrow$CS benchmark in Table \ref{tab:copt_backbone_ablation} show an over 1 \% improvement in mIOU across 19 Cityscapes classes. This indicates that CoPT cooperates with multiple methods and is easily incorporated into state of the art training paradigms. CoPT's training scheme is further ablated in Appendix \ref{sec:supp_training_ablation}.
\par \noindent \textbf{Text Encoder} To test whether CoPT's performance boost is due specifically to CLIP \cite{Radford21} text embeddings or to the use of text embeddings in general, we swap CLIP for Sentence-T5 \cite{Ni22} or Mistral-7B \cite{Jiang23a} and retrain CoPT on GTA$\rightarrow$CS in Table \ref{tab:text_encoder}. CoPT gets state of the art on using both CLIP \cite{Radford21} and Sentence-T5 \cite{Ni22} text embeddings, showing that in general using text embeddings as regularization anchors for pixel feature space benefits the UDA problem. Mistral-7B  \cite{Jiang23a} underperforms because it is trained for next token prediction, leading to worse sentence representations \cite{Jiang23b} and thus worse representations for the domain prompt templates (see Appendix \ref{sec:mistral_emb}).
\par \noindent \textbf{Source vs. Target Pixel Features} In our main experiments CoPT is implemented using class pixel features from the source domain. In the ablation shown in Table \ref{tab:copt_target} we test whether applying CoPT using class features from the target domain is beneficial. To extract class pixel features from the target domain, we use the target pseudo label. CoPT achieves state of the art on GTA$\rightarrow$CS when applied to either source pixel features or target pixel features, showing that adapting the relative distance of class features from each domain to resemble that of text is beneficial. The higher performance on source pixel features is likely due to noise in the target pseudo labels. Due to out-of-memory issues, when applying CoPT to both source and target pixel features we must backpropagate after computing source, which changes the optimization scheme and leads to underperformance.
\section{Discussion}
\label{sec:discussion}
\par Based on the results from our benchmark experiments, CoPT gives the biggest performance boosts on semantic classes with large intra-class variations, such as truck, train and bus in GTA$\rightarrow$CS and wall in Synthia$\rightarrow$CS. These classes are large in reality, leading them to encompass larger swaths of the image than other classes. They also exhibit varying features, such as the vehicles showing ads on the side, or variation due to windows and paint. Trucks, trains and buses in particular have similar appearance as well, but distinct semantic functions, so it is beneficial for the model to learn more distinct embedding spaces for these classes, which CoPT encourages through the use of text.
\par CoPT relies on features from the end of the encoder that are spatially downsampled, which loses detailed information. This can explain why CoPT doesn't improve on the traffic sign class in GTA$\rightarrow$CS, in which different instances of the class have different geometries, requiring the model to use detailed spatial features for accurate boundary segmentation. A naive way to alleviate this problem is to upsample the encoder's features to the original resolution before applying the ground truth mask for each class. However, this leads to out-of-memory errors. Instead, one can upsample crops of the encoder feature to the original resolution and apply the masks to the cropped region, carrying out the rest of the loss as usual. We leave this for future work.

\section{Conclusion}
\label{sec:conclusion}
This work presents the first effort to incorporate vision-language representations into a UDA method for semantic segmentation, a task in which annotations are simultaneously difficult to collect and in dire need. We have introduced a novel covariance-based pixel text loss called CoPT that cooperates with previous UDA methods for semantic segmentation to boost performance, especially on classes with large intra-class variation. CoPT benefits from using domain-agnostic text embeddings based on descriptions of domains generated by LLMs as opposed to hand-crafted ones found in previous methods. CoPT sets a new state of the art based on our experiments on four benchmarks and through our ablations we show that it gives robust improvements to different backbone methods. We hope CoPT inspires future UDA methods to search for novel ways to use domain-agnostic properties of alternative modalities.

\vspace{2em}
\par \noindent \textbf{Acknowledgment}
This work was supported by Electronics and Telecommunications Research Institute (ETRI) grants funded by the Korean government. [24ZR1100, A Study of Hyper-Connected Thinking Internet Technology by autonomous connecting, controlling and evolving ways] and [24ZB1200, Research of Human-centered autonomous intelligence system original technology]

%
%
\bibliographystyle{splncs04}
\bibliography{main}

\begin{thebibliography}{10}
\providecommand{\url}[1]{\texttt{#1}}
\providecommand{\urlprefix}{URL }
\providecommand{\doi}[1]{https://doi.org/#1}

\bibitem{Bose24}
Bose, S., Jha, A., Fini, E., Singha, M., Ricci, E., Banerjee, B.: Stylip: Multi-scale style-conditioned prompt learning for clip-based domain generalization. In: WACV (2024)

\bibitem{Cao23}
Cao, Q., Xu, Z., Chen, Y., Ma, C., Yang, X.: Domain-controlled prompt learning (2023)

\bibitem{Chen22neurips}
Chen, L., Wei, Z., Jin, X., Chen, H., Zheng, M., Chen, K., Jin, Y.: Deliberated domain bridging for domain adaptive semantic segmentation. In: NeurIPS (2022)

\bibitem{Chen23}
Chen, M., Zheng, Z., Yang, Y.: Transferring to real-world layouts: A depth-aware framework for scene adaptation (2023)

\bibitem{Chen23pipa}
Chen, M., Zheng, Z., Yang, Y., Chua, T.S.: Pipa: Pixel- and patch-wise self-supervised learning for domain adaptative semantic segmentation. In: ACM MM (2023)

\bibitem{Chen22}
Chen, R., Rong, Y., Guo, S., Han, J., Sun, F., Xu, T., Huang, W.: Smoothing matters: Momentum transformer for domain adaptive semantic segmentation (2022)

\bibitem{Cordts16}
Cordts, M., Omran, M., Ramos, S., Rehfeld, T., Enzweiler, M., Benenson, R., Franke, U., Roth, S., Schiele, B.: The cityscapes dataset for semantic urban scene understanding. In: CVPR (2016)

\bibitem{Ettedgui22}
Ettedgui, S., Abu-Hussein, S., Giryes, R.: Procst: Boosting semantic segmentation using progressive cyclic style-transfer (2022)

\bibitem{Fahes23}
Fahes, M., Vu, T.H., Bursuc, A., Pérez, P., de~Charette, R.: PØda: Prompt-driven zero-shot domain adaptation. In: ICCV (2023)

\bibitem{Hao23}
Hao, X., Zhang, W., Wu, D., Zhu, F., Li, B.: Dual alignment unsupervised domain adaptation for video-text retrieval. In: CVPR (2023)

\bibitem{Hoyer22cvpr}
Hoyer, L., Dai, D., Gool, L.V.: Daformer: Improving network architectures and training strategies for domain-adaptive semantic segmentation. In: CVPR (2022)

\bibitem{Hoyer22eccv}
Hoyer, L., Dai, D., Gool, L.V.: Hrda: Context-aware high-resolution domain-adaptive semantic segmentation. In: ECCV (2022)

\bibitem{Hoyer23}
Hoyer, L., Dai, D., Wang, H., Gool, L.V.: Mic: Masked image consistency for context-enhanced domain adaptation. In: CVPR (2023)

\bibitem{Hu24}
Hu, X., Wang, K., Zhang, K., Xia, L., Chen, A., Luo, J., Qiao, N., Zeng, X., Sun, M., Kuo, C.H., Sun, Y., Nevatia1, R.: Reclip: Refine contrastive language image pre-training with source free domain adaptation. In: WACV (2024)

\bibitem{Huang23}
Huang, Z., Zhou, A., Lin, Z., Cai, M., Wang, H., Lee, Y.J.: A sentence speaks a thousand images: Domain generalization through distilling clip with language guidance. In: ICCV (2023)

\bibitem{Jiang23a}
Jiang, A.Q., Sablayrolles, A., Mensch, A., Bamford, C., Chaplot, D.S., de~las Casas, D., Bressand, F., Lengyel, G., Lample, G., Saulnier, L., Lavaud, L.R., Lachaux, M.A., Stock, P., Scao, T.L., Lavril, T., Wang, T., Lacroix, T., Sayed, W.E.: Mistral 7b (2023)

\bibitem{Jiang23b}
Jiang, T., Huang, S., Luan, Z., Wang, D., Zhuang, F.: Scaling sentence embeddings with large language models (2023)

\bibitem{Jiang22}
Jiang, Z., Li, Y., Yang, C., Gao, P., Wang, Y., Tai, Y., Wang, C.: Prototypical contrast adaptation for domain adaptive semantic segmentation. In: ECCV (2022)

\bibitem{Jin24}
Jin, S., Jiang, X., Huang, J., Lu, L., Lu, S.: Llms meet vlms: Boost open vocabulary object detection with fine-grained descriptors. In: ICLR (2024)

\bibitem{Kirillov2023}
Kirillov, A., Mintun, E., Ravi, N., Mao, H., Rolland, C., Gustafson, L., Xiao, T., Whitehead, S., Berg, A.C., Lo, W.Y., Doll{\'a}r, P., Girshick, R.: Segment anything. In: ICCV (2023)

\bibitem{Kundu21}
Kundu, J.N., Kulkarni, A., Singh, A., Jampani, V., Babu, R.V.: Generalize then adapt: Source-free domain adaptive semantic segmentation. In: ICCV (2021)

\bibitem{Lai24}
Lai, Z., Bai, H., Zhang, H., Du, X., Shan, J., Yang, Y., Chuah, C.N., Cao, M.: Empowering unsupervised domain adaptation with large-scale pre-trained vision-language models. In: WACV (2024)

\bibitem{Lai23}
Lai, Z., Vesdapunt, N., Zhou, N., Wu, J., Huynh, C.P., Li, X., Fu, K.K., Chuah, C.N.: Padclip: Pseudo-labeling with adaptive debiasing in clip for unsupervised domain adaptation. In: ICCV (2023)

\bibitem{Lee23}
Lee, S., Park, H., Kim, D.U., Kim, J., Boboev, M., Baek, S.: Image-free domain generalization via clip for 3d hand pose estimation. In: WACV (2023)

\bibitem{Li20}
Li, G., Kang, G., Liu, W., Wei, Y., Yang, Y.: Content-consistent matching for domain adaptive semantic segmentation. In: ECCV (2020)

\bibitem{Li22}
Li, J., Wang, Z., Gao, Y., Hu, X.: Exploring high-quality target domain information for unsupervised domain adaptive semantic segmentation. In: ACM MM (2022)

\bibitem{Lu22}
Lu, Y., Luo, Y., Zhang, L., Li, Z., Yang, Y., Xiao, J.: Bidirectional self-training with multiple anisotropic prototypes for domain adaptive semantic segmentation. In: ACM MM (2022)

\bibitem{Min22}
Min, S., Park, N., Kim, S., Park, S., Kim, J.: Grounding visual representations with texts for domain generalization. In: ECCV (2022)

\bibitem{Ni22}
Ni, J., Abrego, G.H., Constant, N., Ma, J., Hall, K., Cer, D., Yang, Y.: Sentence-t5: Scalable sentence encoders from pre-trained text-to-text models. In: ACL (2022)

\bibitem{openai2024gpt4}
OpenAI, :, Achiam, J., Adler, S., Agarwal, S., Ahmad, L., Akkaya, I., Aleman, F.L., Almeida, D., Altenschmidt, J., Altman, S., Anadkat, S., Avila, R., Babuschkin, I., Balaji, S., Balcom, V., Baltescu, P., Bao, H., Bavarian, M., Belgum, J., Bello, I., Berdine, J., Bernadett-Shapiro, G., Berner, C., Bogdonoff, L., Boiko, O., Boyd, M., Brakman, A.L., Brockman, G., Brooks, T., Brundage, M., Button, K., Cai, T., Campbell, R., Cann, A., Carey, B., Carlson, C., Carmichael, R., Chan, B., Chang, C., Chantzis, F., Chen, D., Chen, S., Chen, R., Chen, J., Chen, M., Chess, B., Cho, C., Chu, C., Chung, H.W., Cummings, D., Currier, J., Dai, Y., Decareaux, C., Degry, T., Deutsch, N., Deville, D., Dhar, A., Dohan, D., Dowling, S., Dunning, S., Ecoffet, A., Eleti, A., Eloundou, T., Farhi, D., Fedus, L., Felix, N., Fishman, S.P., Forte, J., Fulford, I., Gao, L., Georges, E., Gibson, C., Goel, V., Gogineni, T., Goh, G., Gontijo-Lopes, R., Gordon, J., Grafstein, M., Gray, S., Greene, R., Gross, J., Gu, S.S., Guo, Y., Hallacy, C.,
  Han, J., Harris, J., He, Y., Heaton, M., Heidecke, J., Hesse, C., Hickey, A., Hickey, W., Hoeschele, P., Houghton, B., Hsu, K., Hu, S., Hu, X., Huizinga, J., Jain, S., Jain, S., Jang, J., Jiang, A., Jiang, R., Jin, H., Jin, D., Jomoto, S., Jonn, B., Jun, H., Kaftan, T., Łukasz Kaiser, Kamali, A., Kanitscheider, I., Keskar, N.S., Khan, T., Kilpatrick, L., Kim, J.W., Kim, C., Kim, Y., Kirchner, J.H., Kiros, J., Knight, M., Kokotajlo, D., Łukasz Kondraciuk, Kondrich, A., Konstantinidis, A., Kosic, K., Krueger, G., Kuo, V., Lampe, M., Lan, I., Lee, T., Leike, J., Leung, J., Levy, D., Li, C.M., Lim, R., Lin, M., Lin, S., Litwin, M., Lopez, T., Lowe, R., Lue, P., Makanju, A., Malfacini, K., Manning, S., Markov, T., Markovski, Y., Martin, B., Mayer, K., Mayne, A., McGrew, B., McKinney, S.M., McLeavey, C., McMillan, P., McNeil, J., Medina, D., Mehta, A., Menick, J., Metz, L., Mishchenko, A., Mishkin, P., Monaco, V., Morikawa, E., Mossing, D., Mu, T., Murati, M., Murk, O., Mély, D., Nair, A., Nakano, R., Nayak,
  R., Neelakantan, A., Ngo, R., Noh, H., Ouyang, L., O'Keefe, C., Pachocki, J., Paino, A., Palermo, J., Pantuliano, A., Parascandolo, G., Parish, J., Parparita, E., Passos, A., Pavlov, M., Peng, A., Perelman, A., de~Avila Belbute~Peres, F., Petrov, M., de~Oliveira~Pinto, H.P., Michael, Pokorny, Pokrass, M., Pong, V.H., Powell, T., Power, A., Power, B., Proehl, E., Puri, R., Radford, A., Rae, J., Ramesh, A., Raymond, C., Real, F., Rimbach, K., Ross, C., Rotsted, B., Roussez, H., Ryder, N., Saltarelli, M., Sanders, T., Santurkar, S., Sastry, G., Schmidt, H., Schnurr, D., Schulman, J., Selsam, D., Sheppard, K., Sherbakov, T., Shieh, J., Shoker, S., Shyam, P., Sidor, S., Sigler, E., Simens, M., Sitkin, J., Slama, K., Sohl, I., Sokolowsky, B., Song, Y., Staudacher, N., Such, F.P., Summers, N., Sutskever, I., Tang, J., Tezak, N., Thompson, M.B., Tillet, P., Tootoonchian, A., Tseng, E., Tuggle, P., Turley, N., Tworek, J., Uribe, J.F.C., Vallone, A., Vijayvergiya, A., Voss, C., Wainwright, C., Wang, J.J., Wang, A.,
  Wang, B., Ward, J., Wei, J., Weinmann, C., Welihinda, A., Welinder, P., Weng, J., Weng, L., Wiethoff, M., Willner, D., Winter, C., Wolrich, S., Wong, H., Workman, L., Wu, S., Wu, J., Wu, M., Xiao, K., Xu, T., Yoo, S., Yu, K., Yuan, Q., Zaremba, W., Zellers, R., Zhang, C., Zhang, M., Zhao, S., Zheng, T., Zhuang, J., Zhuk, W., Zoph, B.: Gpt-4 technical report (2024)

\bibitem{Radford21}
Radford, A., Kim, J.W., Hallacy, C., Ramesh, A., Goh, G., Agarwal, S., Sastry, G., Askell, A., Mishkin, P., Clark, J., Krueger, G., Sutskever, I.: Learning transferable visual models from natural language supervision. In: ICML (2021)

\bibitem{Richter16}
Richter, S.R., Vineet, V., Roth, S., Koltun, V.: Playing for data: Ground truth from computer games. In: ECCV (2016)

\bibitem{Ros16}
Ros, G., Sellart, L., Materzynska, J., Vazquez, D., Lopez, A.M.: The synthia dataset: A large collection of synthetic images for semantic segmentation of urban scenes. In: CVPR (2016)

\bibitem{Sakaridis19}
Sakaridis, C., Dai, D., Gool, L.V.: Guided curriculum model adaptation and uncertainty-aware evaluation for semantic nighttime image segmentation. In: ICCV (2019)

\bibitem{Sakaridis21}
Sakaridis, C., Dai, D., Gool, L.V.: Acdc: The adverse conditions dataset with correspondences for semantic driving scene understanding. In: ICCV (2021)

\bibitem{Truong23}
Truong, T.D., Le, N., Raj, B., Cothren, J., Luu, K.: Fredom: Fairness domain adaptation approach to semantic scene understanding. In: CVPR (2023)

\bibitem{Tsai18}
Tsai, Y.H., Hung, W.C., Schulter, S., Sohn, K., Yang, M.H., Chandraker, M.: Learning to adapt structured output space for semantic segmentation. In: CVPR (2018)

\bibitem{Vidit23}
Vidit, V., Engilberge, M., Salzmann, M.: Clip the gap: A single domain generalization approach for object detection. In: CVPR (2023)

\bibitem{Wang24}
Wang, Z., Zhang, L., Wang, L., Zhu, M.: Landa: Language-guided multi-source domain adaptation (2024)

\bibitem{Wang22}
Wang, Z., Liu, X., Suganuma, M., Okatani, T.: Cross-region domain adaptation for class-level alignment (2022)

\bibitem{Xie23}
Xie, B., Li, S., Li, M., Liu, C.H., Huang, G., Wang, G.: Sepico: Semantic-guided pixel contrast for domain adaptive semantic segmentation. In: IEEE TPAMI (2023)

\bibitem{Xie21}
Xie, E., Wang, W., Yu, Z., Anandkumar, A., Alvarez, J.M., Luo, P.: Segformer: Simple and efficient design for semantic segmentation with transformers. In: NeurIPS (2021)

\bibitem{Zara23}
Zara, G., Conti, A., Roy, S., Lathuiliere, S., Rota, P., Ricci, E.: The unreasonable effectiveness of large language-vision models for source-free video domain adaptation. In: ICCV (2023)

\bibitem{Zhang21}
Zhang, P., Zhang, B., Zhang, T., Chen, D., Wang, Y., Wen, F.: Prototypical pseudo label denoising and target structure learning for domain adaptive semantic segmentation. In: CVPR (2021)

\bibitem{Zheng20}
Zheng, Z., Yang, Y.: Unsupervised scene adaptation with memory regularization in vivo. In: IJCAI (2020)

\end{thebibliography}

\appendix

\section{Additional Related Works}
\label{sec:appendix_related_works}

\subsection{Unsupervised Domain Adaptation for Semantic Segmentation}
\par Self-supervised losses are used by the bulk of previous methods to learn stronger feature representations. In particular, contrastive losses promote learning appearance of objects in the target domain by connecting them to object representations learned in the source domain \cite{Li20, Jiang22, Lu22, Li22, Chen23pipa}. Recently, masked image modeling was shown to give large improvements in performance by Hoyer et al. \cite{Hoyer23}. To recover from errors on rare classes, some works learn priors of the source class distribution and use this to estimate the target class distribution \cite{Truong23, Chen23, Xie23, Hoyer22cvpr}.
\par Domain adaptation methods encompass several data access settings to emulate different real-world applications. In some applications, source data can be proprietary, meaning clients only have access to a source-pretrained model and must adapt it to unlabeled target data. The source-free domain adaptation task was introduced to emulate this setting \cite{Hu24}. Kundu et al. \cite{Kundu21} use data augmentation in the source-free domain adaptation setting. They separate the model training into a stage with access to only the source domain, where they perform data augmentation, then an adaptation stage with access only to unlabeled target data. Image translation, in which source pixels are directly operated on to transform to a target appearance, has been applied to source data and added to the training pipeline \cite{Ettedgui22, Chen22neurips}. In \cite{Chen22}, data from an additional modality, depth, is used. Network architectures and training strategies have been optimized for the UDA segmentation task as well \cite{Hoyer22eccv}.
\par Adversarial learning is attractive because it directly modifies latent features, but can be difficult to optimize and does not currently lead to state of the art performance. Tsai et al. \cite{Tsai18} use adversarial learning in the output/prediction space, modifying the latent features of the segmentation encoder directly to remove domain-specific information. Wang et al. \cite{Wang22} apply an adversarial loss between high and low confidence regions within pseudo-labeled target images. Adversarial learning has recently been revived by Chen et al. \cite{Chen23} who show that when used with self-training it leads transformer models to outperform CNNs on the task. 
\subsection{Domain Adaptation using Vision-Language Embeddings}
\par A common theme in vision-language learning for domain adaptation is to combine text embeddings with image embeddings to imbue semantic details about the target domain into the image representation \cite{Lee23, Vidit23, Fahes23, Wang24}. Min et al. \cite{Min22} train an LSTM-based text generator that outputs explanations for an image classifier's prediction, but this requires ground truth text descriptions of images. In tasks such as video-text retrieval where a joint vision-language embedding space must be learned, Hao et al. \cite{Hao23} perform domain adaptation by maximizing the similarity of target video-text embedding pairs, without the use of CLIP.

\section{Implementation Details}

\subsection{LLM Domain Template Attributes}
\label{sec:supp_llm_attributes}

\par We include the complete set of queries to ChatGPT \cite{openai2024gpt4} whose outputs are used to generate domain descriptors in Table \ref{tab:attributes}. The format of the queries is shown in the top column as ``What makes a <DOMAIN> image look <DOMAIN>?" where the <DOMAIN> can be filled in as synthetic, real, day-time, night-time, snow, fog, and rain depending on the style of the source and target domains. In the GTA$\rightarrow$CS \cite{Cordts16,Richter16} and Synthia$\rightarrow$CS \cite{Ros16} benchmarks, the source domain is synthetic and the target domain is real. In the CS$\rightarrow$DZ \cite{Sakaridis19} benchmark, the source domain is day-time and the target domain is night-time. CS$\rightarrow$ACDC \cite{Sakaridis21} encompasses fog, rain, snow and night-time domains as target and day-time as source. To generate the domain-agnostic text embedding for a class, the class is formatted into the attributes following Equation \ref{eq:llm_query}. The attributes are fed to the frozen text encoder and averaged together.

\subsection{Auxiliary Losses}
\label{sec:auxiliary_losses}

\par In our main experiments we implement CoPT on top of MIC \cite{Hoyer23} because of its state of the art performance on UDA for semantic segmentation. During CoPT training, we also leave the auxiliary losses implemented by default in MIC: pixel-wise cross entropy loss on source, masked pseudo-label self-training using an EMA teacher, ImageNet feature distance regularization, and strongly augmented self-training. The pixel-wise cross entropy loss over source detail is explained in Equation \eqref{eq:ce_loss}. We strongly encourage the reader to refer to Hoyer et al. for details on the other auxiliary losses but give a brief summary of the self-training losses here.
\par For masked pseudo-label self-training, an unlabeled target sample $\mathbf{x}^{t} \in T$ is passed to an EMA-updated teacher model $\mathcal{T}_{\alpha}$ with the same architecture as $\mathcal{E}_{\psi}$ to get pseudo label $\mathbf{y}^{t} = \mathcal{T}_{\alpha}(\mathbf{x}^{t})$. Then a patch in $\mathbf{x}^{t}$ is uniformly sampled and masked to get $\mathbf{x}_{m}^{t}$. The student model's prediction on the masked image is $\mathbf{\hat{y}}_{m}^{t} = \mathcal{E}_{\psi}(\mathbf{x}_{m}^{t})$. The final masked loss is
\begin{align}
    \mathcal{L}_{m} = q^{T}\mathcal{L}_{ce}(\mathbf{\hat{y}}_{m}^{t}, \mathbf{y}^{t})
\label{eq:masked_self_training}
\end{align}
where $q^{t}$ is a segmentation quality estimate. In Hoyer et al. $q^{T}$ is calculated as the ratio of pixels exceeding a threshold of the maximum softmax probability.
\par In strongly augmented self-training, an unlabeled target sample $\mathbf{x}^t$ is strongly augmented to get $\mathbf{\tilde{x}}^{t}$. The original sample is fed to the EMA teacher to get $\mathbf{\hat{y}}^{t} = \mathcal{T}_{\alpha}(\mathbf{x}^{t})$. Then a cross entropy loss is applied to the student's output on the strongly augmented image,
\begin{align}
    \mathcal{L}_{st} = \mathcal{L}_{ce}(\mathcal{E}_{\psi}(\mathbf{\tilde{x}}^{t}), \mathbf{\hat{y}}^{t})
\end{align}

\begin{table}[t]
\begin{minipage}{\textwidth}
\begin{minipage}{0.47\textwidth}
  \caption{Ablation experiment results for memory bank decay are reported as \% mIOU on GTA$\rightarrow$CS over 19 classes.
  }
  \label{tab:membank_ablation}
  \centering
  \begin{tabular}{@{}c@{\hskip 0.5cm}c@{}}
    \toprule
    MemBank Decay & mIOU \\
    \midrule
    0.01 & 70.52 \\
    0.1 & 75.05 \\
    0.5 & \textbf{76.08} \\
  \bottomrule
  \end{tabular}
\end{minipage}
\hspace{0.04\textwidth}
\begin{minipage}{0.47\textwidth}
  \caption{Ablation experiment results for CoPT's training scheme are reported as \% mIOU on GTA$\rightarrow$CS over 19 classes.
  }
  \label{tab:copt_training_ablation}
  \centering
  \begin{tabular}{@{}c@{\hskip 0.5cm}c@{}}
    \toprule
    Method & mIOU \\
    \midrule
    Finetune & 37.34 \\
    Joint Training & \textbf{76.08} \\
  \bottomrule
  \end{tabular} 
\end{minipage}
\end{minipage}
\end{table}
\section{Training Details}
\label{sec:supp_training_details}
\par We re-iterate that in our main experiments, CoPT is implemented on top of the MIC \cite{Hoyer23} code without changes to the default training hyperparameters. This means that the largest image resolution used for training is (1024, 1024), training is done for 40,000 iterations with batch size 2, the image encoder is MiT-B5 \cite{Xie21} and decoder is a DAFormer \cite{Hoyer22cvpr} head. We chose to build on MIC since it is the state of the art for UDA for segmentation and has public code that is simple to run.

\section{Ablation and Analysis Details}

\subsection{Memory Bank Decay} 
\label{sec:supp_membank_ablation}
\par We use a grid search over three values shown in Table \ref{tab:membank_ablation} to choose the decay parameter $\lambda$ in Equation \ref{eq:membank_update} in our pixel feature memory bank. The results indicate that a high decay parameter, where information from previous training iterations is kept for longer, is advantageous to CoPT. Owing to the low batch size, it takes many iterations for the model to see enough samples to give a class enough pixels to draw diverse, class-representative features from. CoPT benefits from diverse pixel features because its goal is to re-orient the average feature of the class in pixel latent space.

\subsection{Joint Training vs. Finetuning}
\label{sec:supp_training_ablation}
\par We evaluate CoPT's optimal training scheme in Table \ref{tab:copt_training_ablation} on the GTA$\rightarrow$CS benchmark. In the joint training row we add CoPT with weight 1 to the other losses during MIC training, and in the finetune row we take a fully trained MIC model and finetune it with just CoPT and source domain cross entropy loss. Our results show that joint training far outperforms finetuning. This implies that learning domain-agnostic features in pixel space using CoPT risks losing discriminative information learned through the self-supervised losses.

\subsection{T-SNE Visualization}
\label{sec:tsne_viz}

\begin{figure}[tb]
  \centering
  \includegraphics[height=5.5cm]{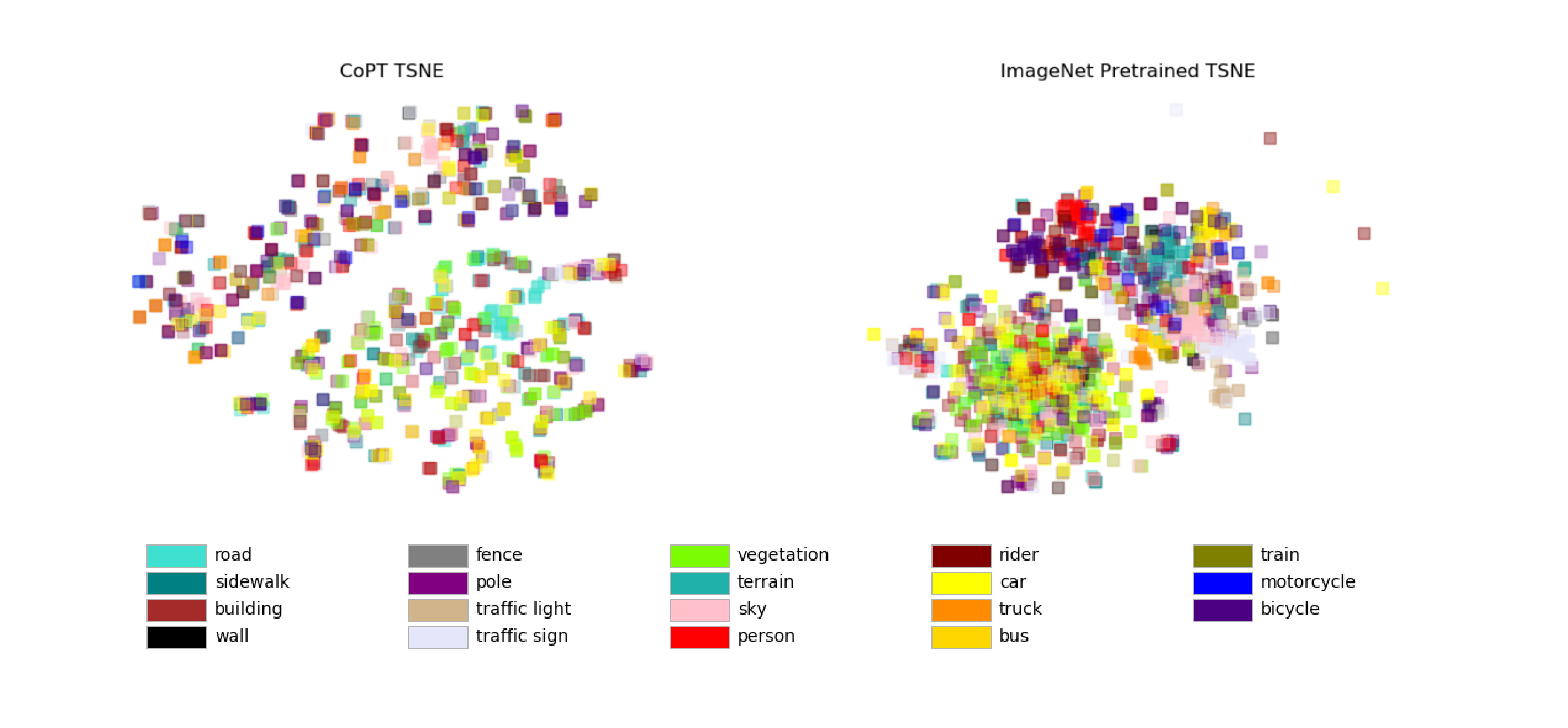}
  \caption{T-SNE visualization on Cityscapes val set of CoPT (left) and ImageNet-pretrained class pixel features (right). Best viewed in color 
  }
  \label{fig:tsne}
\end{figure}

\par Figure \ref{fig:tsne} visualizes t-SNE plots of class pixel features extracted from an MiT-B5 \cite{Xie21} encoder loaded with ImageNet-pretrained weights at the start of training (right) and at the end of training using CoPT (left) implemented on top of MIC \cite{Hoyer23}. Images from Cityscapes val set are fed to the models. CoPT encourages feature embeddings to be more distinct from each other as can be seen by the greater separation between data points. There are fewer outliers in CoPT versus ImageNet e.g. the traffic sign, building and wall classes showing the regularization of the latent space. 

\subsection{Mistral 7B Sentence Embedding}
\label{sec:mistral_emb}

\par When using Mistral 7B \cite{Jiang23a} as CoPT's text embedding method in Section \ref{sec:ablations}, we generate a single text embedding for each prompt by averaging together Mistral's embedding for each word in the prompt. Mistral 7B outputs a single embedding for each word because it was trained for next token prediction and not designed to generate sentence embeddings.

\vspace{2em}
\par \noindent \textbf{Note:} This is the author's accepted manuscript. The final version is published in ECCV 2024 (LNCS vol. 15120), Springer.
\begin{table}[tb]
  \caption{The complete list of ChatGPT \cite{openai2024gpt4} output attributes that are used to generate the domain descriptions as part of the LLM Domain Template process. The domain in the left column is formatted into the ChatGPT Query template.
  Table continued on next page. 
  }
  \label{tab:attributes}
  \centering
  \resizebox{\textwidth}{!}{%
  \begin{tabular}{M{.35\textwidth}M{.65\textwidth}}
    \toprule
    \multicolumn{2}{c}{ChatGPT Query: "What makes a <DOMAIN> image look <DOMAIN>"?}  \\
    \midrule
    DOMAIN & \multicolumn{1}{l}{{Attribute Outputs}} \\
    \midrule
    synthetic & \tiny\renewcommand{\arraystretch}{3}{\begin{enumerate}
        \item lack of realism
        \item unusual colors and lighting
        \item perfect symmetry
        \item repetitive elements
        \item lack of organic variation
        \item tiling and tiling artifacts
        \item overly sharp or soft focus
        \item inconsistent shadows and reflections
        \item distinctive noise patterns
        \item regular patterns and grids
        \item lack of depth and perspective
        \item looks like an inorganic object
        \item an artistic style or stylization
        \item exaggerated features
    \end{enumerate}} \\
    real & {\tiny\renewcommand{\arraystretch}{3}\begin{enumerate}
            \item natural colors and lighting
            \item high resolution
            \item depth and perspective
            \item organic variation
            \item complex textures
            \item authentic shadows and reflections
            \item blurred bokeh background
            \item lens flare and glare
            \item natural poses and expressions
            \item environmental integration
            \item realistic props and objects
            \item accurate reflections in water and glass
            \item natural compression artifacts
            \item weather and atmospheric effects
            \item unscripted candid moments
            \item a tangible sense of scale
    \end{enumerate}} \\
    day-time & {\tiny\renewcommand{\arraystretch}{3}\begin{enumerate}
            \item an abundance of natural light
            \item bright and vibrant colors
            \item soft shadows
            \item warmth in illumination
            \item dynamic range
            \item natural sky colors
            \item distinctive sun position
            \item clear visibility
            \item minimal artificial lighting
            \item lively and active atmosphere
            \item natural textures and patterns
            \item shimmering water bodies
            \item pleasant weather conditions
            \item outdoor shadows
            \item minimal noise and grain
    \end{enumerate}} \\
    night-time & {\tiny\renewcommand{\arraystretch}{3}\begin{enumerate}
            \item low light conditions
            \item dark shadows
            \item diminished color saturation
            \item warm artificial lighting
            \item with contrast between light and dark
            \item point light sources
            \item visible light trails
            \item with glowing skylines
            \item with silhouettes and outlines
            \item with noise and grain
            \item astrophotography elements
            \item long shadows
            \item reflective surfaces
            \item distant atmospheric haze
            \item a sense of mystery and atmosphere
    \end{enumerate}} \\
  \bottomrule
  \end{tabular}}
\end{table}

\begin{table}[t]
   \resizebox{\textwidth}{!}{%
  \begin{tabular}{M{.35\textwidth}M{.65\textwidth}}
    \toprule
    \multicolumn{2}{c}{ChatGPT Query: "What makes a <DOMAIN> image look <DOMAIN>"?}  \\
    \midrule
    DOMAIN & \multicolumn{1}{l}{{Attribute Outputs}} \\
    \midrule
        foggy & {\tiny\renewcommand{\arraystretch}{3}\begin{enumerate}
            \item  soft, diffused light
            \item  hazy atmosphere
            \item  low contrast
            \item  diminished sharpness and clarity
            \item  gradient of opacity
            \item  desaturation of colors
            \item  loss of detail in distance
            \item  diffused light sources
            \item  visible water droplets
            \item  muffled soundscape
            \item  ethereal and dreamlike atmosphere
            \item  silhouettes and silhouetted forms
            \item  moisture on surfaces
            \item  elongated shadows
    \end{enumerate}} \\
        rainy & {\tiny\renewcommand{\arraystretch}{3}\begin{enumerate}
            \item  raindrops on surfaces
            \item  wet and reflective surfaces
            \item  muted colors
            \item  glossy textures
            \item  diminished contrast
            \item  blurred backgrounds
            \item  dynamic water movement
            \item  puddles and reflections
            \item  umbrellas and rain gear
            \item  dramatic sky
            \item  wet vegetation
            \item  water ripples and disturbances
            \item  smeared or distorted lights
            \item  misty atmosphere
    \end{enumerate}} \\
    snowy & {\tiny\renewcommand{\arraystretch}{3}\begin{enumerate}
            \item  white blanket of snow
            \item  soft, diffused light
            \item  cool color palette
            \item  snowflakes in motion
            \item  crystalline texture
            \item  cold, wintry atmosphere
            \item  footprints and tracks
            \item  snow-covered trees and branches
            \item  icy surfaces and frozen water
            \item  winter clothing and gear
            \item  frost and ice crystals
            \item  blurred backgrounds and depth of field
            \item  cold breath and condensation
            \item  winter sports and activities
    \end{enumerate}} \\
    \bottomrule
  \end{tabular}}

\end{table}

\end{document}